%% file: neurips_2026.tex
\documentclass{article}

\usepackage[preprint]{neurips_2026}

\usepackage[utf8]{inputenc} 
\usepackage[T1]{fontenc}    
\usepackage{hyperref}       
\usepackage{url}            
\usepackage{booktabs}       
\usepackage{amsfonts}       
\usepackage{nicefrac}       
\usepackage{microtype}      
\usepackage{xcolor}         

\usepackage{graphicx}
\usepackage{amsmath}
\usepackage{wrapfig} 
\usepackage{natbib}
\usepackage{multirow}
\title{Projection-Volume Fidelity Divergence: Diagnosing and Controlling Optimization Drift in Sparse-View 3D Gaussian Tomography}

\author{%
Yikuang Yuluo$^{1,}$\thanks{Corresponding author.} \quad,
Ao Wang$^{1}$, \quad
Shen Kuan$^{1}$,\quad
Yujie Liu$^{1}$,\quad
Wang Liao$^{1}$, \quad
Ying Chen$^{1}$, \quad\\
Shuangyang Zhong$^{2}$, \quad
Yixing Huang$^{2}$, \quad
Fuquan Wang$^{1}$ \\
$^{1}$Chongqing University, Chongqing, China \\
$^{2}$Peking University Health Science Center, Beijing, China
}

\begin{document}

\maketitle

\begin{abstract}
Sparse-view computed tomography is a severely ill-posed inverse problem, where recent 3D Gaussian Splatting methods offer an efficient explicit representation for tomographic reconstruction. However, we find that projection-domain optimization can be misleading in this setting: the rendered projections may continue to improve while the reconstructed volume deteriorates. We identify this failure mode as Projection-Volume Fidelity Divergence (PVFD), a representation-level optimization drift caused by anisotropic Gaussian deformation and view-specific primitive co-adaptation under sparse Radon constraints. To characterize this behavior, we introduce geometry- and volume-level diagnostics that measure needle-like Gaussian degeneration and the stability of the voxelized density field. Based on these observations, we propose LADES, a ground-truth-free optimization controller for sparse-view Gaussian tomography. LADES combines Linearly Annealed Dropout, which applies strong stochastic masking in early training to disrupt premature primitive co-adaptation and gradually restores full capacity for structural consolidation, with Structure-Aware Early Stopping, which terminates densification according to the saturation of Gaussian population growth rather than validation PSNR. Experiments on sparse-view CT reconstruction show that LADES improves volumetric fidelity, suppresses structural degeneration, and substantially reduces training time while maintaining competitive projection accuracy. These results suggest that robust Gaussian-based tomography requires monitoring and controlling volumetric structure, rather than optimizing projection fit alone.
\end{abstract}

\section{Introduction}

Computed tomography (CT) reconstructs a three-dimensional attenuation field from multi-angle X-ray projections and is widely used in medical diagnosis, industrial inspection, and non-destructive evaluation~\cite{hounsfield1980computed,kak2001principles,de2004distance}. 
When only sparse views are available due to dose, time, safety, or acquisition constraints, CT becomes a severely ill-posed inverse problem. Classical FDK methods are efficient but suffer from streak artifacts, iterative and TV-regularized methods are computationally costly and prior-sensitive~\cite{andersen1984simultaneous,sidky2008image}, and learning-based methods often depend on large paired datasets with limited cross-domain generalization~\cite{lee2018deep,wang2020deep}. 
Recent 3D representations, from NeRF~\cite{mildenhall2021nerf} to 3D Gaussian Splatting (3DGS)~\cite{kerbl20233d}, have enabled projection-supervised reconstruction, with X-ray/CT variants such as Radiative Gaussian Splatting~\cite{cai2024radiative}, R$^2$-Gaussian~\cite{zha2024r}, and DDGS-CT~\cite{gao2024ddgs} showing promising efficiency and fidelity. 
However, in sparse-view CT, projection consistency does not uniquely imply volumetric correctness; the high flexibility of Gaussian primitives may instead over-adapt to sparse projection residuals and compromise the reconstructed 3D density field.

\begin{figure*}[t]
\centering
\includegraphics[width=0.95 \linewidth]{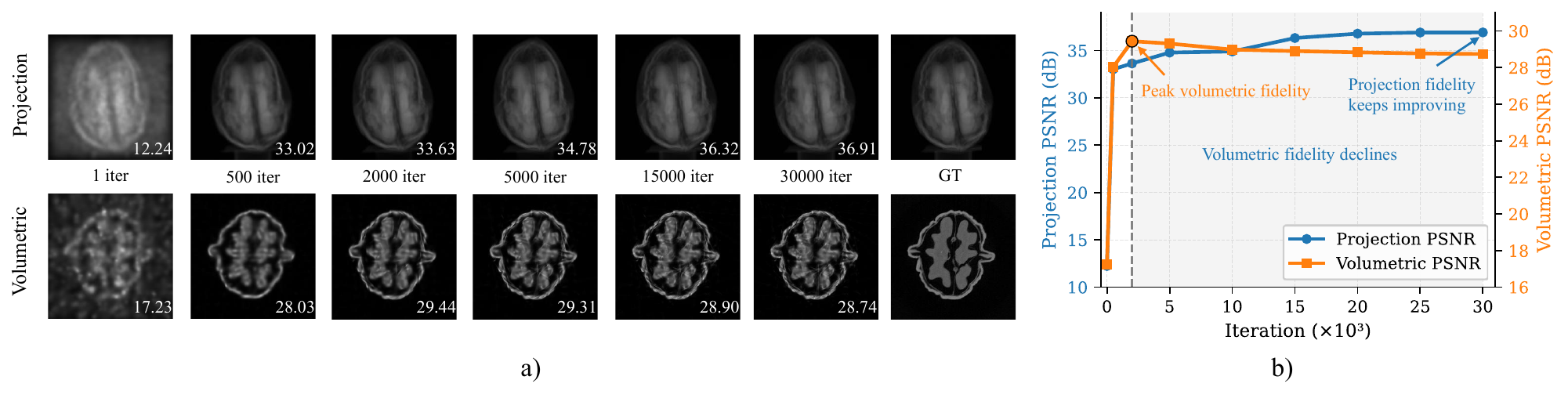} 
\caption{
\textbf{Projection-Volume Fidelity Divergence in sparse-view 3DGS-CT.}
(a) Optimization trajectory of 3DGS-based CT reconstruction. 
The top row shows rendered X-ray projections, and the bottom row shows the corresponding reconstructed axial slice at different training iterations.
(b) Projection-domain PSNR continues to improve, while volumetric fidelity peaks early and then declines.
This mismatch shows that better projection fitting does not necessarily imply better 3D reconstruction under sparse-view constraints, motivating our diagnosis of \emph{Projection-Volume Fidelity Divergence} (PVFD).
}
\label{fig:pvfd}
\end{figure*}

In this work, we identify and characterize this failure mode as \emph{Projection-Volume Fidelity Divergence} (PVFD). 
As illustrated in Fig.~\ref{fig:pvfd}, during projection-supervised 3DGS-CT optimization, the projection-domain PSNR can continue to increase, indicating improved fitting to the observed X-ray measurements. In contrast, the volumetric reconstruction quality reaches its best value at an early stage and then gradually deteriorates. This divergence suggests that late-stage optimization may no longer improve the true 3D structure; instead, it can over-adapt the Gaussian field to sparse-view measurement noise, missing-angle ambiguity, or view-specific residuals. Therefore, projection fidelity is a necessary but insufficient surrogate for volumetric correctness in sparse-view 3DGS-CT.

PVFD is not merely a small fluctuation in reconstruction metrics. 
We observe that it is accompanied by measurable structural degeneration in the Gaussian representation. 
At the primitive level, some Gaussians become highly anisotropic, forming needle-like or plate-like shapes that lack plausible volumetric support but can reduce projection residuals along specific viewing directions. 
At the volume level, the reconstructed density field becomes fragile: perturbing a small subset of primitives can cause disproportionate changes in the voxelized density. 
To quantify these effects, we introduce two diagnostic measures. 
The \emph{Geometric Anisotropy Index} (GAI) measures the shape imbalance of individual Gaussian primitives, while the \emph{Volumetric Co-adaptation Score} (VCS) measures the sensitivity of the reconstructed volume under stochastic primitive perturbations. 
Together, these diagnostics reveal that PVFD reflects a structural overfitting process rather than a harmless late-stage PSNR oscillation.

Existing regularization strategies do not directly address this optimization dynamic. 
Classical CT regularizers are usually defined on voxel grids or image-domain priors, and thus do not explicitly control primitive-level co-adaptation in 3DGS. 
Dropout-based Gaussian regularization can reduce local overfitting by randomly masking primitives, but a fixed dropout rate may restrict the representation throughout training and degrade late-stage detail recovery. 
Conversely, schedules that increase dropout during optimization are designed to regularize late-stage sparse-view rendering, but sparse-view CT requires a different temporal control: strong stochastic regularization is most needed during early structure formation, whereas later optimization should recover the full representational capacity to consolidate density and boundary details. 
This motivates a PVFD-aware training strategy that suppresses early co-adaptation, restores late-stage expressivity, and prevents unnecessary densification after the structure has saturated.

Motivated by this diagnosis, we propose \emph{LADES}, a training control framework for sparse-view 3DGS-CT that combines \emph{Linearly Annealed Dropout} (LAD) and \emph{Structure-Aware Early Stopping} (SAES). 
LAD applies strong random masking at the beginning of training and linearly decays the dropout probability to zero, forcing active primitives to explain stable cross-view structures before full-capacity refinement. 
SAES monitors the Gaussian growth momentum as an intrinsic signal of structural saturation and terminates densification when the primitive population becomes stable. 
Importantly, SAES does not access ground-truth volumes or 3D PSNR during training; ground-truth metrics are used only for post-hoc diagnosis and evaluation. 
By combining early co-adaptation suppression with ground-truth-free structural stopping, LADES mitigates PVFD while preserving the efficiency advantage of 3DGS.

Our contributions are summarized as follows: 1) We identify and characterize Projection-Volume Fidelity Divergence in sparse-view 3DGS-CT, showing that improved projection fitting can coincide with deteriorating volumetric fidelity. 2) We introduce two diagnostic measures, Geometric Anisotropy Index and Volumetric Co-adaptation Score, to quantify primitive-level anisotropy and volume-level fragility associated with PVFD. 3) We propose LADES, a PVFD-aware optimization control framework combining linearly annealed dropout with structure-aware early stopping based on Gaussian growth momentum. 4) We validate the proposed framework across sparse-view settings, reconstruction baselines, and structural diagnostics, demonstrating improved volumetric fidelity and reduced training cost without using ground-truth volumes for training or stopping.

\section{Related Work}

\paragraph{Sparse-view CT and Gaussian-based tomography.}
Sparse-view CT is a highly ill-posed inverse problem due to limited projection measurements.
Classical FBP/FDK methods are efficient but suffer from streak artifacts under undersampling~\cite{feldkamp1984practical}, while iterative methods such as ART, SART, and TV-regularized reconstruction improve stability at the cost of expensive optimization and prior sensitivity~\cite{andersen1984simultaneous,sidky2008image}.
Learning-based methods exploit data-driven priors but often require paired projection-volume data and may generalize poorly across domains~\cite{chen2018learn,zhang2018sparse,wang2020deep,lee2018deep}.
Recent neural representations, including NAF and IntraTomo, optimize continuous attenuation fields from projection supervision~\cite{zha2022naf,zang2021intratomo}, but implicit fields usually require dense sampling and slow optimization.
3D Gaussian Splatting (3DGS) offers an efficient explicit alternative~\cite{kerbl20233d}, and recent X-ray/CT variants such as Radiative Gaussian Splatting, R$^2$-Gaussian, DDGS-CT, TAG-Gaussian, and GR-Gaussian adapt Gaussian primitives to projection-based reconstruction~\cite{cai2024radiative,zha2024r,gao2024ddgs,dai2025tagsplat,yuluo2025gr}.
However, these methods mainly optimize projection consistency and rarely analyze whether improved projection fitting remains aligned with volumetric fidelity under sparse-view constraints.

\paragraph{Regularization and dropout in sparse-view 3DGS.}
Regularization is crucial for sparse-view Gaussian optimization because underconstrained views can induce co-adaptation and view-specific artifacts.
Existing methods use opacity regularization, sparsity constraints, or structure-aware priors to stabilize Gaussian representations~\cite{yuluo2025gr,xiong2023sparsegs}.
Dropout-based methods such as DropoutGS and DropGaussian randomly mask Gaussian primitives to reduce co-adaptation and improve sparse-view rendering generalization~\cite{xu2025dropoutgs,park2025dropgaussian}.
These methods are mainly designed for novel-view synthesis, where the failure mode is appearance degradation in unseen views.
Sparse-view CT differs because supervision is governed by line-integral projection constraints: projection fidelity can continue improving while the underlying 3D density field deteriorates.
Our work identifies this mismatch as PVFD and proposes LADES, which uses early-strong, late-relaxed dropout together with structure-aware densification stopping to control this CT-specific optimization dynamic.

\section{Projection-Volume Fidelity Divergence in 3DGS-CT}
\label{sec:pvfd}
\subsection{Characterizing Projection-Volume Fidelity Divergence}

Projection-supervised optimization treats projection fidelity as a surrogate for volumetric correctness.
This assumption becomes unreliable in sparse-view CT.
As shown in Fig.~\ref{fig:pvfd}, the projection-domain PSNR continues to increase during 3DGS optimization, while the volumetric fidelity reaches an early peak and then gradually declines.
We refer to this mismatch as PVFD.

\begin{figure*}[t]
\centering
\includegraphics[width=0.85\linewidth]{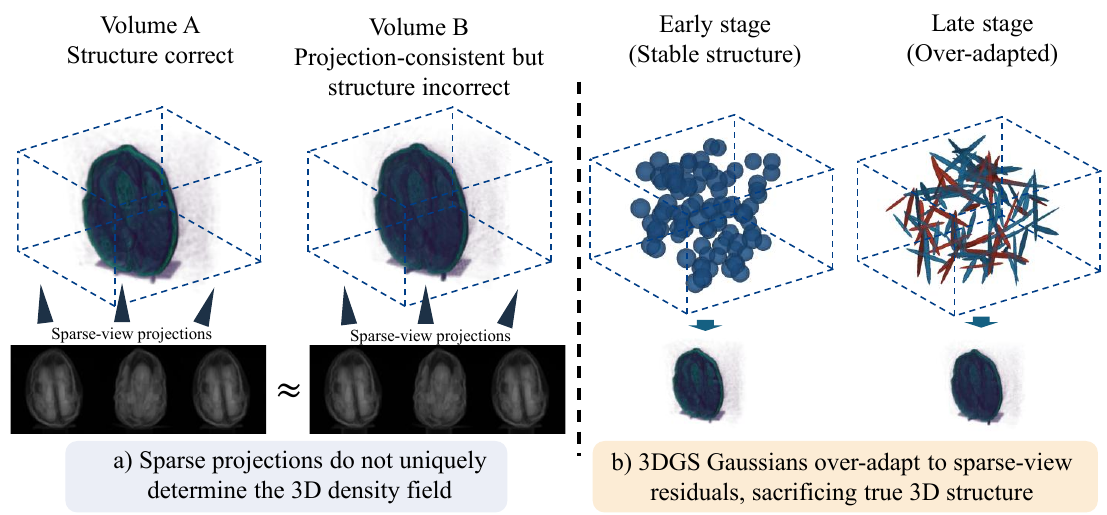}
\caption{
\textbf{Mechanism of Projection-Volume Fidelity Divergence in sparse-view 3DGS-CT.}
(a) Sparse-view CT is underdetermined: different 3D density fields can yield similar sparse projections.
(b) Flexible Gaussian primitives may exploit this ambiguity through anisotropic stretching, view-specific compensation, and fragile co-adapted density patterns.
}
\label{fig:pvfd_mechanism}
\end{figure*}

Fig.~\ref{fig:pvfd_mechanism} illustrates the underlying mechanism.
Sparse-view CT is highly underdetermined: different 3D density fields can explain similar sparse X-ray projections.
Thus, projection consistency is necessary but insufficient for volumetric correctness.
The flexibility of Gaussian primitives further amplifies this ambiguity, since their positions, scales, rotations, and densities provide enough degrees of freedom to fit sparse-view residuals through anisotropic stretching, redundant overlap, or fragile co-adapted density patterns.
Unlike sparse-view novel-view synthesis, where the failure often appears as rendering artifacts in unseen views, sparse-view CT can fail even when the observed projections are fitted increasingly well. Let \(Q_{2D}(t)\) and \(Q_{3D}(t)\) denote the projection-domain and volumetric fidelity at iteration \(t\), respectively.
We define the volumetric peak iteration and PVFD severity as
\begin{equation}
t^\star = \arg\max_t Q_{3D}(t),
\qquad
\Delta_{\mathrm{PVFD}} = Q_{3D}(t^\star) - Q_{3D}(T),
\label{eq:pvfd_severity}
\end{equation}
where \(T\) is the final optimization step.
PVFD occurs when continued optimization improves projection fidelity but degrades volumetric fidelity:
\begin{equation}
Q_{2D}(T) > Q_{2D}(t^\star),
\qquad
Q_{3D}(T) < Q_{3D}(t^\star).
\label{eq:pvfd_condition}
\end{equation}
Importantly, PVFD is not merely a small fluctuation in 3D PSNR.
As shown later, it is accompanied by measurable structural degeneration, including primitive-level anisotropy and volume-level fragility.

\subsection{Quantifying Structural Degeneration}

PVFD is reflected by post-peak degradation in volumetric metrics, but this alone does not explain how the Gaussian representation degenerates.
We therefore introduce two structural diagnostics: primitive-level anisotropy and volume-level fragility.
They measure whether the optimized Gaussian field remains a stable volumetric representation or becomes a set of view-adapted primitives.

\paragraph{Primitive-level anisotropy.}
Let \(s_{i,x}\), \(s_{i,y}\), and \(s_{i,z}\) denote the three principal-axis scales of the \(i\)-th Gaussian primitive.
Under sparse-view projection supervision, some primitives may become needle-like or plate-like to reduce residuals along specific projection directions.
We define the Geometric Anisotropy Index (GAI) as
\begin{equation}
\mathrm{GAI}_i =
\frac{
\max \left(s_{i,x}, s_{i,y}, s_{i,z}\right)
}{
\min \left(s_{i,x}, s_{i,y}, s_{i,z}\right) + \epsilon
},
\label{eq:gai}
\end{equation}
where \(\epsilon\) is a numerical stability constant.
A larger GAI indicates stronger primitive anisotropy.

We summarize field-level degeneration using Global GAI, maximum GAI, and needle ratio:
\begin{equation}
\begin{aligned}
\overline{\mathrm{GAI}} 
&= \frac{1}{N}\sum_{i=1}^{N}\mathrm{GAI}_i, \quad
\mathrm{GAI}_{\max} 
= \max_{1\leq i\leq N}\mathrm{GAI}_i, \\
R_{\mathrm{needle}} 
&= \frac{1}{N}\sum_{i=1}^{N}
\mathbb{I}\!\left(\mathrm{GAI}_i>\tau_{\mathrm{GAI}}\right),
\quad \tau_{\mathrm{GAI}}=50 .
\end{aligned}
\label{eq:gai_statistics}
\end{equation}
Here, \(\overline{\mathrm{GAI}}\) measures average anisotropy, \(\mathrm{GAI}_{\max}\) captures extreme distortion, and \(R_{\mathrm{needle}}\) measures the fraction of highly elongated primitives.

\paragraph{Volume-level fragility.}
Primitive anisotropy captures local geometric distortion, while PVFD may also arise from fragile co-adaptation among primitives.
A stable density field should remain robust when a subset of Gaussians is randomly perturbed.
For each Monte Carlo trial \(k\), we apply a Bernoulli mask and voxelize the retained primitives:
\begin{equation}
m_i^{(k)} \sim \mathrm{Bernoulli}(1-p_{\mathrm{vcs}}),
\qquad
V^{(k)}(x) = V\!\left(x; G \odot m^{(k)}\right).
\label{eq:vcs_mask_volume}
\end{equation}
Given the foreground region \(\Omega_{\mathrm{fg}}\), the Volumetric Co-adaptation Score (VCS) is defined as
\begin{equation}
\mathrm{VCS}
=
\frac{1}{|\Omega_{\mathrm{fg}}|}
\int_{x \in \Omega_{\mathrm{fg}}}
\mathrm{Var}
\left(
\left\{
V^{(k)}(x)
\right\}_{k=1}^{K_{\mathrm{vcs}}}
\right)
dx .
\label{eq:vcs}
\end{equation}
A high VCS indicates that the voxelized density field is sensitive to primitive perturbations, suggesting fragile volumetric co-adaptation.
We use \(p_{\mathrm{vcs}}=0.2\) and \(K_{\mathrm{vcs}}=10\).

\begin{figure*}[t]
\centering
\includegraphics[width=0.95\linewidth]{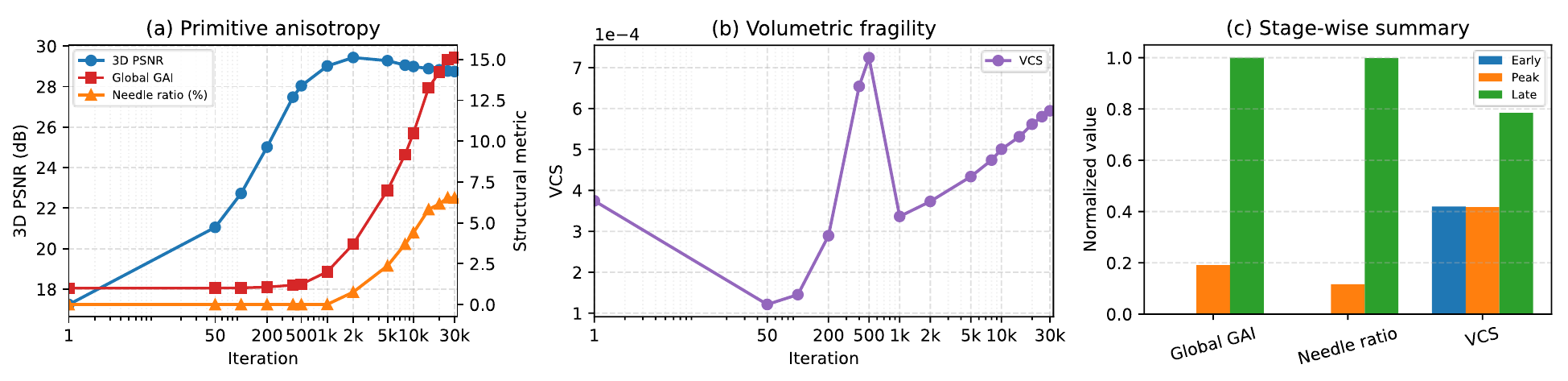}
\caption{
\textbf{Structural diagnostics of PVFD during sparse-view 3DGS-CT optimization.}
(a) After volumetric fidelity peaks, Global GAI and needle ratio continue to increase, indicating progressive primitive-level anisotropy.
(b) VCS shows a sustained upward trend in the post-peak PVFD region, suggesting increasing volume-level fragility.
(c) Stage-wise comparison summarizes the accumulation of structural degeneration.
Together, these diagnostics show that PVFD is not merely a PSNR fluctuation, but a structural degeneration process of the Gaussian representation.
}
\label{fig:structural_diagnostics}
\end{figure*}

\paragraph{Observation.}
Fig.~\ref{fig:structural_diagnostics} shows that structural degeneration becomes evident after the volumetric peak.
The needle ratio increases from \(0.76\%\) at 2k iterations to \(2.37\%\) at 5k, \(4.39\%\) at 10k, and \(6.53\%\) at 30k, indicating rapid accumulation of highly elongated primitives.
Meanwhile, VCS exhibits a sustained upward trend in the post-peak region, showing that the voxelized density field becomes increasingly fragile.
These results indicate that PVFD corresponds to structural degeneration, rather than a harmless late-stage fluctuation in 3D PSNR.

\subsection{Alleviating PVFD with LADES}

The diagnostics above suggest that PVFD is driven by two coupled effects.
First, during early structure formation, Gaussian primitives can rapidly develop view-specific co-adaptation under sparse projection supervision.
Second, after the main volumetric structure has saturated, continued densification may introduce anisotropic and fragile primitives that further reduce projection residuals but degrade the underlying density field.
Motivated by this diagnosis, we propose \emph{LADES}, a PVFD-aware training control framework that combines \emph{Linearly Annealed Dropout} (LAD) and \emph{Structure-Aware Early Stopping of Densification} (SAES).
LAD suppresses early co-adaptation by applying strong stochastic masking during structure formation, while SAES detects structural saturation from Gaussian population dynamics and switches the model to fixed-topology refinement.

\paragraph{Linearly annealed dropout.}
At iteration \(t\), we sample a Bernoulli mask for each Gaussian primitive:
\begin{equation}
m_i^{(t)} \sim \mathrm{Bernoulli}(1-p_t),
\label{eq:lad_mask}
\end{equation}
where \(m_i^{(t)}=1\) means that the \(i\)-th primitive is retained for the current forward and backward pass.
The projection loss is then computed using the masked Gaussian set:
\begin{equation}
\mathcal{L}_{\mathrm{LAD}}(t)
=
\sum_{v\in\mathcal{V}}
\left\|
\hat{P}_v(G_t \odot m^{(t)}) - P_v
\right\|_1
+
\lambda_{\mathrm{SSIM}}
\mathcal{L}_{\mathrm{SSIM}}
\left(
\hat{P}_v(G_t \odot m^{(t)}), P_v
\right).
\label{eq:lad_loss}
\end{equation}

Unlike fixed dropout, which restricts representation capacity throughout training, LAD applies the strongest stochastic constraint at the beginning and gradually restores the full Gaussian field:
\begin{equation}
p_t =
\begin{cases}
p_0 \max\left(0, 1-\frac{t}{T_{\mathrm{anneal}}}\right), & t < t_s, \\
0, & t \geq t_s,
\end{cases}
\label{eq:lad_schedule}
\end{equation}
where \(p_0\) is the initial dropout probability, \(T_{\mathrm{anneal}}\) is the annealing horizon, and \(t_s\) is the SAES trigger iteration.
In our implementation, \(p_0=0.9\) and \(T_{\mathrm{anneal}}=30{,}000\). The key intuition is that early high dropout prevents primitives from relying on a fixed set of neighboring Gaussians to fit view-specific residuals.
Only primitives that remain useful under different random masks receive stable gradient updates.
As training progresses, the dropout probability decreases, allowing the model to recover full capacity for local boundary refinement and density consolidation.
This is different from increasing dropout schedules designed for late-stage regularization in sparse-view rendering: in sparse-view CT, the critical period is early structure formation, where non-physical co-adaptation first emerges.

\paragraph{Structure-aware early stopping of densification.}
LAD improves the structure formation process, but continued clone/split/prune operations after structural saturation can still introduce redundant or anisotropic primitives.
We therefore introduce SAES to stop densification using only intrinsic Gaussian population dynamics.
Let \(N_j\) denote the number of active Gaussians at the \(j\)-th monitoring step.
We record \(N_j\) every \(\Delta\) iterations and compute the normalized population growth rate:
\begin{equation}
g_j =
\frac{
N_j - N_{j-1}
}{
\max(N_{j-1},1)
}.
\label{eq:growth_rate}
\end{equation}
To reduce stochastic fluctuations caused by clone, split, and prune operations, we average the growth rate over a sliding window:
\begin{equation}
\bar{g}_j =
\frac{1}{W-1}
\sum_{q=j-W+2}^{j}
g_q .
\label{eq:growth_average}
\end{equation}
SAES is triggered when the averaged growth rate falls below a threshold:
\begin{equation}
t_s =
\min
\left\{
t_j \mid t_j \geq t_{\min},\ \bar{g}_j < \tau
\right\}.
\label{eq:saes_trigger}
\end{equation}
In our implementation, \(t_{\min}=2{,}000\), \(\Delta=100\), \(W=5\), and \(\tau=10^{-3}\). Once SAES is triggered, we stop all topology-changing operations and switch to fixed-topology refinement:
\begin{equation}
T_{\mathrm{densify}} = t_s,
\qquad
T_{\mathrm{final}} = 2t_s .
\label{eq:saes_final}
\end{equation}
At this transition, dropout is disabled and the learning rates of geometry-related parameters are cooled down:
\begin{equation}
\eta_{\mathrm{xyz}} \leftarrow \gamma_{\mathrm{cool}}\eta_{\mathrm{xyz}},
\qquad
\eta_{\mathrm{scale}} \leftarrow \gamma_{\mathrm{cool}}\eta_{\mathrm{scale}},
\qquad
\gamma_{\mathrm{cool}}=0.2 .
\label{eq:lr_cooling}
\end{equation}
We also reset the first- and second-order Adam moments of the position and scaling parameters to remove the optimization inertia accumulated during the growth stage.
This stabilizes the transition from topology expansion to fixed-topology refinement. Importantly, SAES is not a ground-truth-based early stopping rule.
It does not access ground-truth volumes, 3D PSNR, 3D SSIM, or any test-time reconstruction metric during training.
The volumetric metrics shown in Fig.~\ref{fig:pvfd} and Fig.~\ref{fig:structural_diagnostics} are used only for post-hoc diagnosis and evaluation.

\section{Experiments}

\subsection{Experimental Setup}

\paragraph{Dataset and protocols.}
We evaluate all methods on the real X-ray CT dataset from the Finnish Inverse Problems Society (FIPS)~\cite{FIPS_CT_dataset}, using three representative objects: \emph{walnut}, \emph{pinecone}, and \emph{seashell}.
For each object, we uniformly sample \(10\), \(20\), and \(25\) views from the original acquisition to construct sparse-view inputs.
Unless otherwise specified, projections have resolution \(560\times560\), reconstructed volumes are evaluated at \(256^3\), and all density values are normalized to \([0,1]\).

\paragraph{Baselines.}
We compare LADES with classical CT reconstruction methods and recent Gaussian-based reconstruction methods.
Classical baselines include FDK~\cite{feldkamp1984practical} and SART~\cite{andersen1984simultaneous}.
Gaussian-based baselines include TAG-Gaussian~\cite{dai2025tagsplat}, GR-Gaussian~\cite{yuluo2025gr}, and R\(^2\)-Gaussian~\cite{zha2024r}.
For controlled analysis, we further evaluate fixed dropout, LAD-only, SAES-only, and the full LADES framework under the same R\(^2\)-Gaussian backbone.

\paragraph{Implementation.}
All 3DGS-based methods are implemented in PyTorch and trained on a single NVIDIA RTX 4090 GPU.
The maximum initial training budget is \(30{,}000\) iterations.
For LADES, we use \(p_0=0.9\), \(T_{\mathrm{anneal}}=30{,}000\), \(t_{\min}=2{,}000\), monitoring interval \(\Delta=100\), window size \(W=5\), and growth threshold \(\tau=10^{-3}\).
When SAES is triggered at \(t_s\), densification is stopped and fixed-topology refinement continues until \(2t_s\).
SAES never accesses ground-truth volumes, 3D PSNR, 3D SSIM, or any test-time reconstruction metric during training.

\paragraph{Metrics.}
We report projection-domain PSNR/SSIM, volumetric PSNR/SSIM, training time, and final iteration count.
To evaluate PVFD, we report the post-peak degradation \(\Delta_{\mathrm{PVFD}}\) defined in Sec.~\ref{sec:pvfd}.
To evaluate structural stability, we report Global GAI, needle ratio, and VCS.
For dropout-schedule ablations, SAES is disabled and all schedule-only variants are trained with the same \(30{,}000\)-iteration budget.

\begin{figure*}[!t]
\centering
\includegraphics[width=0.95\linewidth]{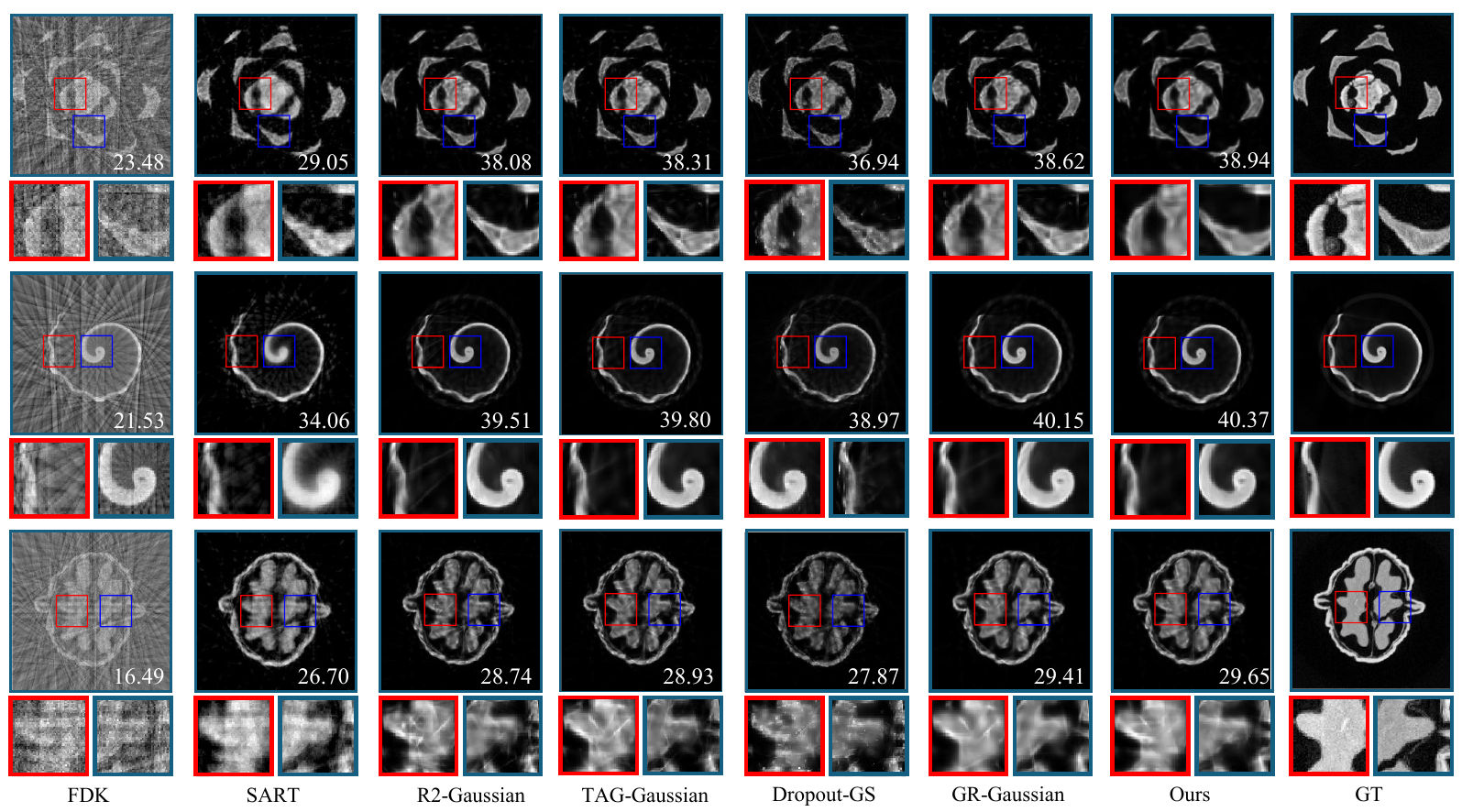}
\caption{
Visual comparison on FIPS under the 25-view sparse setting.
}
\label{fig:main_results_visual}
\end{figure*}

\subsection{Main Results and PVFD Mitigation}

\begin{wraptable}{!r}{0.48\textwidth}
\centering
\scriptsize
\setlength{\tabcolsep}{3pt}
\renewcommand{\arraystretch}{0.95}
\caption{
Quantitative comparison on FIPS under the 25-view sparse setting.
Results are averaged over three real scanned objects.
The \(\pm\) term will denote repeated-run variation when available.
}
\begin{minipage}{\linewidth}
\centering
\begin{tabular}{@{}lccc@{}}
\toprule
\textbf{Method} 
& \textbf{3D PSNR$\uparrow$} 
& \textbf{3D SSIM$\uparrow$} 
& \textbf{Time}
\\
\midrule
FDK
& \(20.50\pm0.03\) & \(0.152\pm0.002\) & 0.5 min\\
SART
& \(29.94\pm0.06\) & \(0.789\pm0.004\) & 1.7 min\\
R$^2$-Gaussian
& \(35.45\pm0.02\) & \(0.842\pm0.003\) & 10.3 min\\
TAG-Gaussian
& \(35.68\pm0.05\) & \(0.844\pm0.002\) & 10.2 min\\
Dropout-GS
& \(34.59\pm0.04\) & \(0.843\pm0.005\) & 19.5 min\\
GR-Gaussian
& \(36.06\pm0.03\) & \(0.844\pm0.003\) & 11.4 min\\
\textbf{LADES (Ours)}
& $36.33\pm0.02$ & $0.853\pm0.002$ & 3.7 min \\
\bottomrule
\end{tabular}
\end{minipage}
\label{tab:main_results_25view}
\end{wraptable}

\paragraph{Quantitative comparison.}
Table~\ref{tab:main_results_25view} reports the 25-view reconstruction results on the FIPS dataset.
We compare LADES with classical CT reconstruction methods, including FDK and SART, and Gaussian-based reconstruction methods, including TAG-Gaussian, GR-Gaussian, R$^2$-Gaussian, and Dropout-GS.
LADES achieves the best average 3D PSNR among all compared methods.
Compared with R$^2$-Gaussian, LADES improves the average 3D PSNR from \(35.45\) dB to \(36.33\) dB.
It also outperforms GR-Gaussian, the strongest baseline in this comparison, by \(0.27\) dB on average.
This suggests that PVFD-aware training control improves volumetric reconstruction quality beyond simply fitting sparse projections.



\paragraph{Visual comparison.}
Figure~\ref{fig:main_results_visual} shows representative reconstructions for the same 25-view setting.
The three rows correspond to different FIPS objects, and all methods are evaluated with the same slice locations and zoomed regions.
Classical methods exhibit severe streak artifacts and blurred structures under sparse-view sampling.
Gaussian-based methods substantially improve global reconstruction quality, but R$^2$-Gaussian, TAG-Gaussian, Dropout-GS, and GR-Gaussian still show residual boundary distortion or local density diffusion in the zoomed regions.
LADES produces sharper local structures and more coherent boundaries, which is consistent with its higher average 3D PSNR in Table~\ref{tab:main_results_25view}.

\begin{figure*}[t]
\centering
\includegraphics[width=0.95\linewidth]{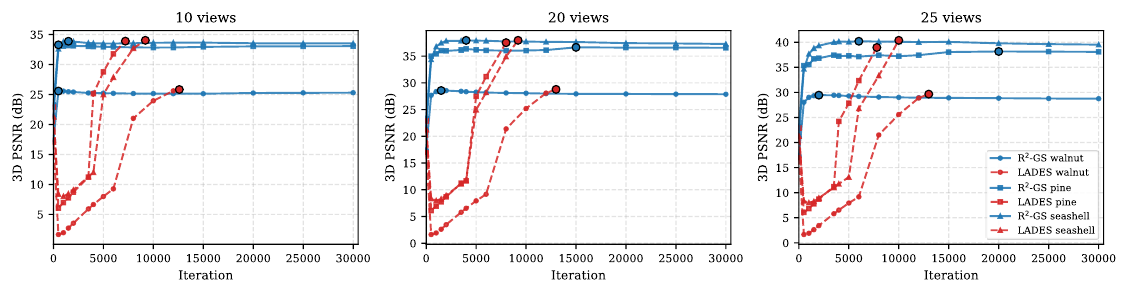}
\caption{
PVFD mitigation dynamics across sparse-view settings.
We compare 3D PSNR trajectories of R$^2$-GS and LADES under \(10\), \(20\), and \(25\) views on FIPS.
}
\label{fig:pvfd_mitigation_curves}
\end{figure*}

\paragraph{PVFD mitigation dynamics.}
Final reconstruction metrics alone do not show whether LADES mitigates the optimization dynamics behind PVFD.
Figure~\ref{fig:pvfd_mitigation_curves} compares 3D PSNR trajectories of R$^2$-GS and LADES across different sparse-view settings.
R$^2$-GS reaches an early volumetric peak and then suffers post-peak degradation when trained to the fixed iteration budget.
By contrast, LADES follows the SAES-controlled trajectory, stops densification once Gaussian population growth saturates, and enters fixed-topology refinement.

Table~\ref{tab:pvfd_structural_dynamics} further quantifies PVFD severity and structural stability.
Under 25 views, R$^2$-GS has \(\Delta_{\mathrm{PVFD}}=0.484\), Global GAI \(=17.09\), needle ratio \(=7.74\%\), and VCS \(=3.72\times10^{-4}\).
LADES reduces the observed post-peak degradation to \(0.000\) along its SAES-terminated trajectory, while also lowering Global GAI to \(6.52\), needle ratio to \(0.77\%\), and VCS to \(6.55\times10^{-5}\).
Fixed dropout reduces some structural indicators but suffers severe volumetric underfitting, showing that structural diagnostics must be interpreted jointly with final 3D fidelity.
Overall, LADES mitigates PVFD at both the metric and representation levels.

\begin{table*}[t]
\centering
\scriptsize
\setlength{\tabcolsep}{3pt}
\renewcommand{\arraystretch}{0.95}
\caption{
PVFD severity and structural stability on FIPS.
\(\Delta_{\mathrm{PVFD}}\) measures post-peak 3D PSNR degradation along the training trajectory.
Global GAI, needle ratio, and VCS quantify primitive-level anisotropy and volume-level fragility. N/A means PVFD does not occur.
}
\begin{tabular}{l|c|cccccc}
\toprule
\textbf{Method}
& \textbf{Views}
& \textbf{3D PSNR$\uparrow$}
& \textbf{3D SSIM$\uparrow$}
& \(\boldsymbol{\Delta_{\mathrm{PVFD}}\downarrow}\)
& \textbf{Global GAI$\downarrow$}
& \textbf{Needle (\%)$\downarrow$}
& \textbf{VCS$\downarrow$} \\
\midrule
R$^2$-GS
& 10 & 30.63 & 0.799  & 0.287 & 28.49 & 13.93 & \(4.38{\times}10^{-4}\) \\
R$^2$-GS
& 20 & 33.91 & 0.832 & 0.480 & 19.91 & 9.52 & \(4.09{\times}10^{-4}\) \\
R$^2$-GS
& 25 & 35.45 & 0.842 & 0.484 & 17.09 & 7.74 & \(3.72{\times}10^{-4}\) \\

\textbf{LADES}
& 10 & \textbf{31.26} & 0.810 & N/A & \textbf{7.34} & \textbf{1.51} & \textbf{$6.31{\times}10^{-5}$} \\
\textbf{LADES}
& 20 & \textbf{34.76} & 0.843 & N/A & \textbf{6.80} & \textbf{0.92} & \textbf{$6.67{\times}10^{-5}$} \\
\textbf{LADES}
& 25 & \textbf{36.33} & 0.853 & N/A & \textbf{6.52} & \textbf{0.77} & \textbf{$6.55{\times}10^{-5}$} \\
\bottomrule
\end{tabular}
\label{tab:pvfd_structural_dynamics}
\end{table*}

\begin{figure*}[t]
\centering
\includegraphics[width=0.95\linewidth]{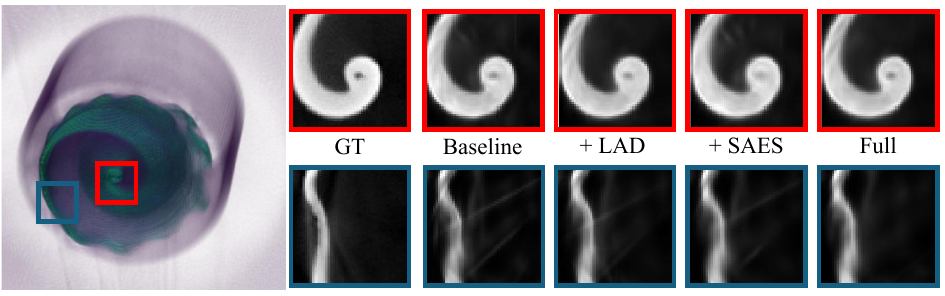}
\caption{
Visual ablation of LAD and SAES under the 25-view FIPS setting.
}
\label{fig:ablation_visual}
\end{figure*}

\subsection{Ablation on Dropout Schedules and SAES}

We ablate the two components of LADES under the 25-view FIPS setting.
All variants use the same R$^2$-GS backbone, projection loss, and densification rule.
Schedule-only variants are trained with the same \(30{,}000\)-iteration budget and SAES disabled, isolating the effect of dropout scheduling.
SAES-only disables LAD and uses Gaussian population growth to control densification.

Table~\ref{tab:schedule_saes_ablation} shows that naive stochastic regularization is insufficient.
Fixed dropout lowers VCS compared with R$^2$-GS, but its 3D PSNR drops sharply from \(35.45\) dB to \(30.62\) dB and its \(\Delta_{\mathrm{PVFD}}\) remains high, indicating over-regularization and volumetric underfitting.
LAD-only and SAES-only isolate two complementary effects: LAD controls early view-specific co-adaptation, while SAES prevents redundant late-stage densification after Gaussian population growth saturates.
Full LADES combines both mechanisms, achieving the best 3D PSNR, the smallest observed post-peak degradation, and the lowest VCS.

Figure~\ref{fig:ablation_visual} provides a visual ablation under the same sparse-view setting.
The baseline suffers from projection-driven structural distortion, while fixed dropout suppresses some co-adaptation but produces over-smoothed or underfitted density structures.
LAD-only improves early structure formation but can still retain late-stage redundancy.
SAES-only reduces unnecessary densification but lacks early stochastic control.
Full LADES produces the most coherent local structure and sharpest boundaries, consistent with the quantitative improvements in Table~\ref{tab:schedule_saes_ablation}.

\begin{table*}[!htbp]
\centering
\scriptsize
\setlength{\tabcolsep}{3pt}
\renewcommand{\arraystretch}{0.95}
\caption{
Ablation of LAD and SAES under the 25-view FIPS setting.
Schedule-only variants are trained with SAES disabled and the same \(30{,}000\)-iteration budget. N/A means PVFD does not occur.
}
\begin{tabular}{@{}lcccccc@{}}
\toprule
\textbf{Variant}
& \textbf{Schedule}
& \textbf{SAES}
& \textbf{3D PSNR$\uparrow$}
& \textbf{3D SSIM$\uparrow$}
& \(\boldsymbol{\Delta_{\mathrm{PVFD}}\downarrow}\)
& \textbf{VCS$\downarrow$} \\
\midrule
Baseline R$^2$-GS
& N/A & \(\times\)
& 35.45 & 0.842 & 0.484 & \(3.72{\times}10^{-4}\) \\
Fixed Dropout
& Constant & \(\times\)
& 30.62 & 0.805 & 0.607 & \(1.99{\times}10^{-4}\) \\
LAD-only
& \(p_t\downarrow\) & \(\times\)
& 36.03 & 0.851 & N/A &$5.87{\times}10^{-5}$ \\
SAES-only
& None & \(\checkmark\)
& 36.12 & 0.850 & N/A &$1.54{\times}10^{-4}$ \\
\textbf{Full LADES}
& \(p_t\downarrow\) & \(\checkmark\)
& \textbf{36.33} & 0.853 & N/A & $6.55{\times}10^{-5}$ \\
\bottomrule
\end{tabular}

\label{tab:schedule_saes_ablation}
\end{table*}

\paragraph{Parameter sensitivity.}
We further evaluate the sensitivity of LADES to its two key hyper-parameters: the initial dropout probability \(p_0\) in LAD and the growth threshold/window size \((\tau,W)\) in SAES.
As shown in Table~\ref{tab:param_sensitivity}, moderate variations around the default setting lead to similar reconstruction quality and PVFD mitigation.
A too-small \(p_0\) weakens early co-adaptation suppression, while a too-large \(p_0\) over-regularizes the Gaussian field.
Similarly, SAES is stable around \(\tau=10^{-3}\) and \(W=5\), indicating that the stopping behavior is not the result of a narrowly tuned threshold.


\begin{table*}[!htbp]
\centering
\scriptsize
\setlength{\tabcolsep}{3pt}
\renewcommand{\arraystretch}{0.95}
\caption{
Parameter sensitivity of LADES under the 25-view FIPS setting.
}
\begin{tabular}{@{}lccc|lccc@{}}
\toprule
\multicolumn{4}{c|}{\textbf{LAD initial dropout}} 
& \multicolumn{4}{c}{\textbf{SAES stopping criterion}} \\
\cmidrule(lr){1-4}\cmidrule(lr){5-8}
\textbf{Setting}
& \textbf{PSNR$\uparrow$}
& \textbf{SSIM$\uparrow$}
& \textbf{VCS$\downarrow$}
& \textbf{Setting}
& \textbf{PSNR$\uparrow$}
& \textbf{SSIM$\uparrow$}
& \textbf{VCS$\downarrow$} \\
\midrule
\(p_0=0.85\)
& \( 36.05\) 
& \( 0.850\) 
& \( 9.80{\times}10^{-5}\)
& $\tau=5\times10^{-4}, W=5$
& \( 36.18\) 
& \( 0.851\) 
&  $8.20{\times}10^{-5}$ \\

\(p_0=0.90\) 
& \(\textbf{36.33}\) 
& \(\textbf{0.853}\) 
& $6.55\times10^{-5}$
& $\tau=10^{-3}, W=5$
& \(\textbf{36.33}\) 
& \(\textbf{0.853}\) 
& $6.55\times10^{-5}$ \\

\(p_0=0.95\)
& \( 36.10\) 
& \( 0.851\) 
& \( 7.40{\times}10^{-5}\)
& $\tau=2\times10^{-3}, W=5$
& \( 36.11\) 
& \( 0.850\) 
&  $7.10\times10^{-5}$ \\
\bottomrule
\end{tabular}
\label{tab:param_sensitivity}
\end{table*}

\section{Limitations and Conclusion}

We studied PVFD in sparse-view 3DGS-CT, where projection fidelity continues to improve while volumetric fidelity degrades after an early peak. We showed that PVFD is not merely a metric fluctuation, but is accompanied by structural degeneration of the Gaussian representation, quantified by primitive anisotropy and volumetric fragility. To mitigate this failure mode, we proposed LADES, combining linearly annealed dropout with structure-aware early stopping of densification. LADES suppresses early view-specific co-adaptation and stops redundant late-stage densification using only intrinsic Gaussian population dynamics. Experiments on real FIPS CT data demonstrate improved volumetric fidelity and reduced PVFD without ground-truth-based stopping. Our current evaluation is limited to sparse-view CT on FIPS with primarily Gaussian-based reconstruction backbones. Generalization to other scanner geometries, noise levels, and clinical acquisition protocols requires further validation. We do not claim clinical deployment readiness.

\bibliography{my.bib}

\newpage

\appendix

\section*{Appendix Overview}
\label{app:overview}
\addcontentsline{toc}{section}{Appendix Overview}

The appendix is organized into five sections. Table~\ref{tab:appendix_overview} 
provides a quick reference for the content of each section, intended to 
help reviewers locate specific results and analyses.

\begin{table}[h]
\centering
\small
\renewcommand{\arraystretch}{1.25}
\begin{tabular}{p{0.5cm} p{4.5cm} p{8.5cm}}
\toprule
\textbf{\S} & \textbf{Section} & \textbf{Content} \\
\midrule
A & Notation and Symbol Table 
  & Complete notation reference grouped by semantic category: 
    Gaussian representation, projection/reconstruction, PVFD characterization, 
    structural diagnostics, LAD, and SAES. 
    (Appendix~\ref{app:notation}) \\
\addlinespace[2pt]
B & Formal Analysis of PVFD 
  & Theoretical perspective on why PVFD arises in sparse-view 3DGS-CT, 
    including null-space analysis of the sparse Radon transform, an informal 
    proposition on projection-equivalent volumetric ambiguity, and 
    interpretations of LAD and SAES. 
    (Appendix~\ref{app:formal_analysis}) \\
\addlinespace[2pt]
C & Implementation Details 
  & Backbone configuration, optimizer and learning rates, data preprocessing, 
    LAD/SAES implementation including Adam-moment reset, hardware and runtime, 
    and reproducibility notes (random seeds, code release). 
    (Appendix~\ref{app:implementation}) \\
\addlinespace[2pt]
D & Extended Experimental Results 
  & Per-object quantitative results across all view counts, per-object 
    structural diagnostics, SAES trigger iterations, training time analysis, 
    and correlation between diagnostics and PVFD severity. 
    (Appendix~\ref{app:extended_results}) \\
\addlinespace[2pt]
E & Diagnostic Metrics: Validation and Sensitivity 
  & Implementation details and validation of GAI/VCS, including correlation 
    with PVFD severity, sensitivity to threshold and sampling parameters, 
    pairwise correlation across diagnostics, and discussion of alternative 
    designs. 
    (Appendix~\ref{app:diagnostic_validation}) \\
\bottomrule
\end{tabular}
\caption{Appendix overview. The appendix is structured to support 
independent reading: each section is self-contained and cross-references 
relevant equations, tables, and figures from both the main paper and 
other appendix sections.}
\label{tab:appendix_overview}
\end{table}

\section{Notation and Symbol Table}
\label{app:notation}

For convenience, we summarize the notation used throughout the paper. 
Symbols are grouped by semantic category: 
(i) Gaussian representation (Table~\ref{tab:notation-gaussian}), 
(ii) projection and volumetric reconstruction (Table~\ref{tab:notation-projection}), 
(iii) PVFD characterization (Table~\ref{tab:notation-pvfd}), 
(iv) structural diagnostics (Table~\ref{tab:notation-diagnostics}), 
(v) Linearly Annealed Dropout (Table~\ref{tab:notation-lad}), and 
(vi) Structure-Aware Early Stopping (Table~\ref{tab:notation-saes}).

\begin{table}[h]
\centering
\caption{Notation: Gaussian representation.}
\label{tab:notation-gaussian}
\small
\renewcommand{\arraystretch}{1.15}
\begin{tabular}{p{2.6cm} p{2.0cm} p{8.4cm}}
\toprule
\textbf{Symbol} & \textbf{Domain} & \textbf{Description} \\
\midrule
$G_t $ & --- & Set of 3D Gaussian primitives at iteration $t$ \\
$N_t$ & $\mathbb{N}$ & Number of active Gaussian primitives at iteration $t$ \\
$s_{i,x}, s_{i,y}, s_{i,z}$ & $\mathbb{R}_{>0}$ & Principal-axis scales of the $i$-th Gaussian primitive \\
$V(\mathbf{x};\,G)$ & $\mathbb{R}_{\ge 0}$ & Voxelized density at location $\mathbf{x}$ rendered from Gaussian set $\mathcal{G}$ \\
$\Omega_{\text{fg}}$ & $\subset \mathbb{R}^3$ & Foreground region of the reconstructed volume \\
\bottomrule
\end{tabular}
\end{table}

\begin{table}[h]
\centering
\caption{Notation: projection and volumetric reconstruction.}
\label{tab:notation-projection}
\small
\renewcommand{\arraystretch}{1.15}
\begin{tabular}{p{2.6cm} p{2.0cm} p{8.4cm}}
\toprule
\textbf{Symbol} & \textbf{Domain} & \textbf{Description} \\
\midrule
$\mathcal{V}$ & --- & Set of available sparse-view projection angles \\
$P_v$ & $\mathbb{R}^{H \times W}$ & Ground-truth X-ray projection at view $v$ \\
$\hat{P}_v(G)$ & $\mathbb{R}^{H \times W}$ & Rendered projection at view $v$ from Gaussian set $\mathcal{G}$ \\
$\mathcal{L}_{\text{SSIM}}$ & $\mathbb{R}_{\ge 0}$ & SSIM-based loss between rendered and ground-truth projections \\
$\lambda_{\text{SSIM}}$ & $\mathbb{R}_{\ge 0}$ & Weight of the SSIM loss term \\
$Q_{2D}(t)$ & $\mathbb{R}$ & Projection-domain reconstruction fidelity (e.g., 2D PSNR) at iteration $t$ \\
$Q_{3D}(t)$ & $\mathbb{R}$ & Volumetric reconstruction fidelity (e.g., 3D PSNR) at iteration $t$ \\
\bottomrule
\end{tabular}
\end{table}

\begin{table}[h]
\centering
\caption{Notation: PVFD characterization.}
\label{tab:notation-pvfd}
\small
\renewcommand{\arraystretch}{1.15}
\begin{tabular}{p{2.6cm} p{2.0cm} p{8.4cm}}
\toprule
\textbf{Symbol} & \textbf{Domain} & \textbf{Description} \\
\midrule
$t^{\star}$ & $\mathbb{N}$ & Volumetric peak iteration, $t^{\star} = \arg\max_t Q_{3D}(t)$ \\
$T$ & $\mathbb{N}$ & Final optimization iteration \\
$\Delta_{\text{PVFD}}$ & $\mathbb{R}_{\ge 0}$ & Post-peak volumetric degradation, $\Delta_{\text{PVFD}} = Q_{3D}(t^{\star}) - Q_{3D}(T)$ \\
\bottomrule
\end{tabular}
\end{table}

\begin{table}[h]
\centering
\caption{Notation: structural diagnostics (GAI and VCS).}
\label{tab:notation-diagnostics}
\small
\renewcommand{\arraystretch}{1.15}
\begin{tabular}{p{2.6cm} p{2.0cm} p{8.4cm}}
\toprule
\textbf{Symbol} & \textbf{Domain} & \textbf{Description} \\
\midrule
$\text{GAI}_i$ & $\mathbb{R}_{\ge 1}$ & Geometric Anisotropy Index of the $i$-th primitive \\
$\overline{\text{GAI}}$ & $\mathbb{R}_{\ge 1}$ & Field-level mean Geometric Anisotropy Index (also referred to as Global GAI) \\
$\text{GAI}_{\max}$ & $\mathbb{R}_{\ge 1}$ & Maximum Geometric Anisotropy Index in the field \\
$R_{\text{needle}}$ & $[0, 1]$ & Fraction of needle-like primitives with $\text{GAI}_i > \tau_{\text{GAI}}$ \\
$\tau_{\text{GAI}}$ & $\mathbb{R}_{>0}$ & Threshold for needle-like primitives ($\tau_{\text{GAI}} = 50$) \\
$\epsilon$ & $\mathbb{R}_{>0}$ & Numerical stability constant in GAI denominator ($\epsilon = 10^{-8}$) \\
$\text{VCS}$ & $\mathbb{R}_{\ge 0}$ & Volumetric Co-adaptation Score \\
$m_i^{(k)}$ & $\{0, 1\}$ & Bernoulli mask for primitive $i$ in VCS Monte Carlo trial $k$ \\
$p_{\text{vcs}}$ & $[0, 1]$ & Bernoulli drop probability for VCS perturbation ($p_{\text{vcs}} = 0.2$) \\
$K_{\text{vcs}}$ & $\mathbb{N}$ & Number of Monte Carlo trials for VCS ($K_{\text{vcs}} = 10$) \\
$V^{(k)}(\mathbf{x})$ & $\mathbb{R}_{\ge 0}$ & Voxelized density under the $k$-th perturbed Gaussian set \\
$\mathbb{I}(\cdot)$ & $\{0, 1\}$ & Indicator function \\
\bottomrule
\end{tabular}
\end{table}

\begin{table}[h]
\centering
\caption{Notation: Linearly Annealed Dropout (LAD).}
\label{tab:notation-lad}
\small
\renewcommand{\arraystretch}{1.15}
\begin{tabular}{p{2.6cm} p{2.0cm} p{8.4cm}}
\toprule
\textbf{Symbol} & \textbf{Domain} & \textbf{Description} \\
\midrule
$m_i^{(t)}$ & $\{0, 1\}$ & Bernoulli mask for primitive $i$ at training iteration $t$ \\
$p_t$ & $[0, 1]$ & Dropout probability at iteration $t$ \\
$p_0$ & $[0, 1]$ & Initial dropout probability ($p_0 = 0.9$) \\
$T_{\text{anneal}}$ & $\mathbb{N}$ & Annealing horizon for LAD ($T_{\text{anneal}} = 30{,}000$) \\
$\mathcal{L}_{\text{LAD}}(t)$ & $\mathbb{R}_{\ge 0}$ & Projection loss with LAD masking at iteration $t$ \\
$\odot$ & --- & Element-wise (Hadamard) product over primitives \\
\bottomrule
\end{tabular}
\end{table}

\begin{table}[h]
\centering
\caption{Notation: Structure-Aware Early Stopping (SAES).}
\label{tab:notation-saes}
\small
\renewcommand{\arraystretch}{1.15}
\begin{tabular}{p{2.6cm} p{2.0cm} p{8.4cm}}
\toprule
\textbf{Symbol} & \textbf{Domain} & \textbf{Description} \\
\midrule
$N_j$ & $\mathbb{N}$ & Active Gaussian count at the $j$-th monitoring step \\
$g_j$ & $\mathbb{R}$ & Normalized population growth rate at step $j$ \\
$\bar{g}_j$ & $\mathbb{R}$ & Sliding-window-averaged growth rate \\
$\Delta$ & $\mathbb{N}$ & Monitoring interval in iterations ($\Delta = 100$) \\
$W$ & $\mathbb{N}$ & Sliding window size for growth-rate smoothing ($W = 5$) \\
$\tau$ & $\mathbb{R}_{>0}$ & Growth-rate threshold for SAES trigger ($\tau = 10^{-3}$) \\
$t_{\min}$ & $\mathbb{N}$ & Minimum iteration before SAES can trigger ($t_{\min} = 2{,}000$) \\
$t_s$ & $\mathbb{N}$ & SAES trigger iteration \\
$T_{\text{densify}}$ & $\mathbb{N}$ & Final iteration of densification, $T_{\text{densify}} = t_s$ \\
$T_{\text{final}}$ & $\mathbb{N}$ & Final training iteration, $T_{\text{final}} = 2t_s$ \\
$\eta_{xyz}, \eta_{\text{scale}}$ & $\mathbb{R}_{>0}$ & Learning rates for position and scale parameters \\
$\gamma_{\text{cool}}$ & $(0, 1]$ & Learning-rate cooldown factor at SAES trigger ($\gamma_{\text{cool}} = 0.2$) \\
\bottomrule
\end{tabular}
\end{table}

\section{Formal Analysis of PVFD}
\label{app:formal_analysis}

In this section, we provide a formal perspective on Projection-Volume 
Fidelity Divergence (PVFD). Our goal is not to establish a fully rigorous 
theorem, but to clarify why PVFD is an intrinsic consequence of 
sparse-view projection supervision combined with the high geometric 
flexibility of Gaussian primitives. We organize the analysis into four 
parts: (B.1) the null space of the sparse-view Radon transform, (B.2) 
an informal proposition on projection-equivalent volumetric ambiguity, 
(B.3) a bias--variance interpretation of LAD, and (B.4) a structural 
interpretation of why Gaussian population growth rate signals structural 
saturation.

\subsection{Sparse-View Radon Transform and Its Null Space}
\label{app:null_space}

Let $V: \mathbb{R}^3 \to \mathbb{R}_{\ge 0}$ denote a continuous 
attenuation field, and let $R_v$ denote the Radon (line-integral) 
operator along projection direction $v$. Given $K$ projection views 
$\mathcal{V} = \{v_1, \ldots, v_K\}$, the sparse-view Radon transform 
is the concatenated operator
\begin{equation}
R_{\mathcal{V}}(V) = \big( R_{v_1}(V), \ldots, R_{v_K}(V) \big) 
\in \mathbb{R}^{K \times h \times w}.
\end{equation}
For dense-view CT (i.e., $K \to \infty$), classical results show that 
$R$ is injective on suitable function spaces and $V$ can be recovered 
from $R(V)$ up to bounded error. For sparse-view CT, however, $K$ is 
finite (in our experiments $K \in \{10, 20, 25\}$), and $R_{\mathcal{V}}$ 
is a strongly underdetermined linear operator. Its null space
\begin{equation}
\mathcal{N}(R_{\mathcal{V}}) 
= \big\{ \delta V \;\big|\; R_{\mathcal{V}}(\delta V) = 0 \big\}
\end{equation}
is non-trivial and contains an infinite family of volumetric 
perturbations that are invisible to all $K$ projection measurements. 
Consequently, projection consistency alone cannot uniquely determine 
the underlying 3D density field: any solution $V$ admits an equivalence 
class $\{ V + \delta V \mid \delta V \in \mathcal{N}(R_{\mathcal{V}}) \}$ 
that fits the observed projections equally well. This null-space 
ambiguity is the fundamental source of ill-posedness in sparse-view CT 
and motivates the use of structural priors or regularizers in classical 
reconstruction methods~\cite{sidky2008image}.

\subsection{An Informal Proposition: Projection-Equivalent Volumetric Ambiguity in 3DGS-CT}
\label{app:proposition}

The null-space ambiguity discussed above is well known for voxel-based 
representations. We now argue that this ambiguity is amplified in 
3DGS-CT due to the anisotropic flexibility of Gaussian primitives. 
Concretely, the Gaussian field can exploit $\mathcal{N}(R_{\mathcal{V}})$ 
through configurations that produce nearly identical projections but 
substantially different volumetric structures.

Let $G$ denote a 3D Gaussian field, where each primitive 
$g_i$ has free anisotropic scales $(s_{i,x}, s_{i,y}, s_{i,z}) \in 
\mathbb{R}_{>0}^3$. Let $V(\cdot;G)$ be the voxelization 
operator and $R_{\mathcal{V}}$ the sparse-view Radon transform with 
$K$ views. If primitives are allowed to be sufficiently anisotropic 
(i.e., the ratio $\max_a s_{i,a} / \min_a s_{i,a}$ is unbounded), 
then for any $\epsilon > 0$, there exist two Gaussian configurations 
$G_A \ne G_B$ such that
\begin{equation}
\big\| R_{\mathcal{V}}\big(V(\cdot;G_A)\big) 
- R_{\mathcal{V}}\big(V(\cdot;G_B)\big) \big\|_2 < \epsilon,
\end{equation}
yet
\begin{equation}
\big\| V(\cdot;G_A) - V(\cdot;G_B) \big\|_2 
\;\gg\; \epsilon.
\end{equation}

\paragraph{Construction sketch.} 
The proposition can be illustrated by a simple two-Gaussian construction. 
Consider a sparse-view setting with $K$ projection directions 
$\{v_1, \ldots, v_K\}$. For any direction $v^{\perp} \notin \mathcal{V}$ 
that lies in the angular gap between adjacent measured views, one can 
place a needle-like Gaussian $g^{\text{needle}}$ whose major axis aligns 
with $v^{\perp}$. Because the line integral of an elongated Gaussian 
along its major axis is highly localized in a thin region orthogonal 
to $v^{\perp}$, $g^{\text{needle}}$ contributes negligibly to $R_{v_k}$ 
for projection directions sufficiently far from $v^{\perp}$. Adding or 
removing such needle primitives produces volumetric configurations 
$G_A$ and $G_B$ whose projections differ only by $O(\epsilon)$, but 
whose voxelized volumes can differ by $O(1)$ in mass at the location 
of $g^{\text{needle}}$. This construction extends to arbitrarily many 
non-measured directions in $S^2 \setminus \mathcal{V}$, yielding a 
high-dimensional family of projection-equivalent but volumetrically 
distinct Gaussian configurations.

\paragraph{Implication.} 
Proposition formalizes the intuition behind 
Fig.~2 in the main paper. Under sparse-view supervision, the 
projection loss gradient does not penalize the formation of 
needle-aligned anisotropic primitives that occupy the angular null 
space of $R_{\mathcal{V}}$. As optimization continues, gradient updates 
that further reduce projection residuals along measured directions 
can simultaneously increase volumetric error along unmeasured 
directions. This is precisely the optimization dynamic that 
$\Delta_{\text{PVFD}}$, $\overline{\text{GAI}}$, and $R_{\text{needle}}$ 
quantify in the main paper.

\subsection{Bias--Variance Interpretation of Linearly Annealed Dropout}
\label{app:lad_analysis}

We next provide an informal interpretation of why LAD mitigates the 
ambiguity described above. Consider an ensemble view of the Gaussian 
field. At iteration $t$, applying a Bernoulli mask $m^{(t)}$ with 
drop probability $p_t$ effectively trains a sub-field 
$G_t \odot m^{(t)}$ with expected size 
$(1-p_t) N_t$. Over many iterations, the optimization minimizes the 
expected projection loss
\begin{equation}
\mathbb{E}_{m^{(t)}} 
\big[ \mathcal{L}_{\text{proj}}(G_t \odot m^{(t)}) \big],
\end{equation}
which decomposes into a term penalizing the average prediction and a 
variance term penalizing predictions that depend strongly on specific 
co-active subsets of primitives. The variance term implicitly suppresses 
fragile co-adaptation: a needle-aligned primitive whose projection 
contribution is canceled by a specific neighbor produces high prediction 
variance under random masking, and is therefore disfavored by the 
expected loss.

\paragraph{Why annealing.} 
A constant dropout rate biases the optimization toward representations 
that are robust under masking but may lack the capacity to fit fine 
boundary details, which is consistent with the volumetric underfitting 
of fixed dropout observed in Table~3 of the main paper. By annealing 
$p_t$ to zero, LAD applies the strongest variance penalty during early 
structure formation---when co-adaptation is most likely to emerge---and 
gradually restores full representational capacity for late-stage 
refinement. This is fundamentally different from increasing dropout 
schedules used in sparse-view rendering, where the failure mode is 
late-stage view-specific overfitting rather than early structural 
co-adaptation.

\subsection{Why Gaussian Population Growth Saturates at Structural Convergence}
\label{app:saes_analysis}

Finally, we explain why Gaussian population growth rate $\bar{g}_j$ 
serves as an intrinsic, ground-truth-free signal of structural 
saturation. In the standard 3DGS densification rule, clone and split 
operations are triggered when the position gradient magnitude of a 
primitive exceeds a threshold:
\begin{equation}
\| \nabla_{\mathbf{x}_i} \mathcal{L}_{\text{proj}} \|_2 > \tau_{\text{grad}}.
\end{equation}
During early optimization, the projection loss exhibits large-scale 
structural residuals, and gradient magnitudes are concentrated on a 
small number of primitives near under-represented structural regions. 
This produces frequent densification events and a high growth rate.

As the main volumetric structure forms, residuals become spatially 
diffuse: the remaining errors are distributed across many primitives 
in roughly equal magnitudes, and few primitives individually exceed 
$\tau_{\text{grad}}$. Consequently, the densification rate naturally 
decays. Empirically, we observe that the moving-averaged growth rate 
$\bar{g}_j$ falls below $10^{-3}$ shortly after the volumetric peak 
identified in Fig.~1, suggesting that growth-rate saturation is 
correlated with structural convergence.

\paragraph{Why this is preferable to projection-based stopping.} 
A natural alternative is to stop training when projection PSNR plateaus. 
However, as established by Proposition, projection 
PSNR can continue improving even after volumetric structure has 
converged, by exploiting null-space ambiguity. Stopping based on 
projection PSNR therefore tends to over-train and accumulate 
needle-like primitives. In contrast, the population growth rate 
reflects the internal optimization dynamics of densification rather 
than the external projection fit, making it a more reliable 
ground-truth-free proxy for structural convergence.

\subsection{Summary}
\label{app:formal_summary}

Together, the four arguments above provide a coherent explanation 
of PVFD: (i) sparse-view projection supervision admits a non-trivial 
null space, (ii) anisotropic Gaussian primitives can exploit this 
null space to form projection-equivalent but volumetrically inconsistent 
configurations, (iii) early-strong stochastic masking suppresses the 
formation of such fragile configurations through an implicit 
variance penalty, and (iv) Gaussian population growth rate provides 
an intrinsic signal of structural convergence that is decoupled 
from projection fit. We emphasize that this analysis is intended as 
a conceptual framework rather than a fully formal proof, and we leave 
a complete operator-theoretic treatment of 3DGS-CT inverse problems 
to future work.

\section{Implementation Details}
\label{app:implementation}

This section provides implementation details that complement the 
experimental setup in Section~4 of the main paper. We describe 
(C.1) the 3DGS-CT backbone configuration, (C.2) the data preprocessing 
pipeline, (C.3) the implementation of Linearly Annealed Dropout, 
(C.4) the implementation of Structure-Aware Early Stopping including 
the Adam-moment reset, (C.5) baseline configurations, (C.6) hardware 
and runtime characteristics, and (C.7) random seeds and reproducibility 
notes.

\subsection{3DGS-CT Backbone Configuration}
\label{app:backbone}

We build LADES on top of the $\text{R}^2$-Gaussian backbone~\cite{r2_gaussian}, 
which represents the attenuation field as a set of anisotropic 3D 
Gaussians and renders X-ray projections via a differentiable line-integral 
rasterizer. We retain the original densification, opacity activation, 
and rasterization implementation, and only modify the components 
relevant to LAD and SAES.

\paragraph{Initialization.} 
The Gaussian field is initialized from a coarse FDK reconstruction of 
the input sparse-view projections. Initial primitives are sampled from 
voxels with attenuation values above a foreground threshold 
($0.05 \times \text{maximum attenuation}$). The initial number of 
primitives is approximately $N_0 = 5{,}000$ across all FIPS objects, 
with isotropic scales initialized to a uniform prior derived from the 
inter-primitive distance.

\paragraph{Optimizer and learning rates.} 
All Gaussian parameters are optimized with Adam 
with $\beta_1 = 0.9$, $\beta_2 = 0.999$, and $\epsilon_{\text{Adam}} = 10^{-15}$. 
The per-parameter learning rates are:
\begin{itemize}
\item Position $\eta_{xyz}$: $1.6 \times 10^{-4}$ (decayed exponentially to $1.6 \times 10^{-6}$)
\item Scale $\eta_{\text{scale}}$: $5 \times 10^{-3}$
\item Rotation $\eta_{\text{rot}}$: $1 \times 10^{-3}$
\item Opacity $\eta_{\alpha}$: $5 \times 10^{-2}$
\item Density $\eta_{\rho}$: $5 \times 10^{-2}$
\end{itemize}

\paragraph{Densification.} 
Following the standard 3DGS densification rule~\cite{kerbl20233d}, 
clone and split operations are triggered when the accumulated position 
gradient norm exceeds $\tau_{\text{grad}} = 2 \times 10^{-4}$. 
Densification is performed every $100$ iterations, starting from 
iteration $500$. Primitives with opacity below $0.005$ are pruned at 
each densification step. These backbone settings are kept identical 
across all baselines and LADES variants for fair comparison.

\paragraph{Loss function.} 
The base projection loss combines an $L_1$ term and an SSIM term:
\begin{equation}
\mathcal{L}_{\text{proj}}(G) = \sum_{v \in \mathcal{V}} 
\Big[ \|\hat{P}_v(G) - P_v\|_1 
+ \lambda_{\text{SSIM}} \mathcal{L}_{\text{SSIM}}\big(\hat{P}_v(G), P_v\big) \Big],
\end{equation}
with $\lambda_{\text{SSIM}} = 0.2$. When LAD is active, the projection 
loss is computed using the masked Gaussian set $G_t \odot m^{(t)}$ 
as defined in Eq.~(8) of the main paper.

\subsection{Data Preprocessing}
\label{app:preprocessing}

We use three real X-ray CT scans from the FIPS dataset~\cite{FIPS_CT_dataset}: 
walnut, pinecone, and seashell. For each object, the original acquisition 
contains several hundred projections; we uniformly subsample 
$K \in \{10, 20, 25\}$ views to construct the sparse-view input.

\paragraph{Projection normalization.} 
Each projection $P_v$ is converted from raw intensity to attenuation 
via the Beer--Lambert law:
\begin{equation}
P_v = -\log\frac{I_v}{I_0},
\end{equation}
where $I_0$ is the air-scan intensity. The resulting attenuation 
projections are normalized to the range $[0, 1]$ using the global 
maximum across all measured views.

\paragraph{Volume normalization and resolution.} 
The reconstructed volume is evaluated on a $256^3$ grid with attenuation 
values normalized to $[0, 1]$ using the same global scaling factor as 
the projections. Projections have resolution $560 \times 560$.

\paragraph{Geometry.} 
The FIPS scans use a cone-beam geometry. The source-to-detector and 
source-to-object distances, the detector pixel pitch, and the rotation 
axis are read from the scan metadata provided with the dataset and used 
unchanged in the Radon-transform forward model.

\subsection{Linearly Annealed Dropout: Implementation}
\label{app:lad_impl}

\paragraph{Mask sampling granularity.} 
At each training iteration $t$, we sample a single Bernoulli mask 
$m^{(t)} \in \{0,1\}^{N_t}$ shared across all $K$ projection views in 
the current iteration. We do not sample a separate mask per view, since 
per-view masking would weaken the cross-view consistency signal that 
LAD is designed to enforce.

\paragraph{Dropout schedule.} 
The dropout probability follows the linear schedule defined in Eq.~(9) 
of the main paper:
\begin{equation}
p_t = 
\begin{cases}
p_0 \cdot \max\!\left(0,\; 1 - \dfrac{t}{T_{\text{anneal}}}\right), & t < t_s \\[6pt]
0, & t \ge t_s
\end{cases}
\end{equation}
with $p_0 = 0.9$ and $T_{\text{anneal}} = 30{,}000$. Dropout is fully 
disabled once SAES is triggered at $t_s$, even if $t_s < T_{\text{anneal}}$, 
to allow unconstrained late-stage refinement.

\paragraph{Interaction with densification.} 
Dropped primitives at iteration $t$ do not contribute to the rendered 
projections and therefore do not accumulate position gradients in that 
iteration. To prevent dropout from biasing the densification statistics, 
we use the standard 3DGS gradient-accumulation buffer that averages 
gradients over the densification interval ($100$ iterations), so that 
each primitive contributes to densification proportionally to its 
expected retention probability $1 - p_t$.

\paragraph{Computational overhead.} 
LAD adds one Bernoulli sampling per iteration ($O(N_t)$) and one 
element-wise mask multiplication on the Gaussian parameter tensor 
before rasterization. The empirical overhead is below $1\%$ of the 
total per-iteration time and is dominated by GPU rasterization.

\subsection{Structure-Aware Early Stopping: Implementation}
\label{app:saes_impl}

\paragraph{Population monitoring.} 
We record the active Gaussian count $N_j$ every $\Delta = 100$ 
iterations starting from iteration $0$. The normalized growth rate 
and the sliding-window-averaged growth rate are computed as in 
Eqs.~(10)--(11) of the main paper, with window size $W = 5$ 
(i.e., the average is taken over the past $500$ iterations of training).

\paragraph{Trigger condition.} 
SAES is triggered at the first monitoring step satisfying both 
$t_j \ge t_{\min}$ and $\bar{g}_j < \tau$, with $t_{\min} = 2{,}000$ 
and $\tau = 10^{-3}$. The minimum-iteration constraint $t_{\min}$ 
prevents premature triggering during the early initialization phase, 
when the population can briefly stabilize before structural growth 
begins.

\paragraph{Post-trigger transition.} 
Once SAES is triggered at iteration $t_s$, we apply the following 
operations within a single iteration:
\begin{enumerate}
\item Disable all topology-changing operations (clone, split, prune).
\item Disable LAD masking by setting $p_t = 0$ for all subsequent iterations.
\item Cool down the learning rates of geometry parameters: 
$\eta_{xyz} \leftarrow \gamma_{\text{cool}} \cdot \eta_{xyz}$ and 
$\eta_{\text{scale}} \leftarrow \gamma_{\text{cool}} \cdot \eta_{\text{scale}}$, 
with $\gamma_{\text{cool}} = 0.2$. Rotation, opacity, and density 
learning rates are unchanged.
\item Reset Adam first- and second-moment buffers ($m_t^{\text{Adam}}$ 
and $v_t^{\text{Adam}}$) for the position and scale parameters only. 
This removes optimization inertia accumulated during the growth stage 
and stabilizes the transition to fixed-topology refinement. Adam 
moments for rotation, opacity, and density are kept unchanged to 
preserve their slower, more stable optimization trajectories.
\item Set the final iteration to $T_{\text{final}} = 2 t_s$.
\end{enumerate}

\paragraph{What SAES does not access.} 
We emphasize that SAES never accesses the ground-truth volume, 
3D PSNR, 3D SSIM, GAI, VCS, or any test-time reconstruction metric. 
The only signals used to determine $t_s$ are (i) the active Gaussian 
count $N_j$, which is an intrinsic property of the model, and 
(ii) the iteration index $t_j$. The volumetric metrics shown in 
Figs.~1, 3, and 5 of the main paper are computed offline for diagnosis 
and evaluation only.

\subsection{Baseline Configurations}
\label{app:baselines}

For fair comparison, all 3DGS-based baselines use the same backbone, 
projection loss, and densification rule as LADES. Method-specific 
hyperparameters follow the original publications.

\paragraph{Classical baselines.} 
FDK is implemented using the 
\texttt{tigre} toolkit with cone-beam geometry. SART
is run for $50$ iterations with relaxation factor $0.5$ and 
non-negativity projection.

\paragraph{Gaussian-based baselines.} 
$\text{R}^2$-Gaussian, TAG-Gaussian, GR-Gaussian, and Dropout-GS are 
all trained for $30{,}000$ iterations under the same FIPS sparse-view 
protocol. Dropout-GS uses a constant per-primitive dropout rate of 
$0.1$ following~\cite{xu2025dropoutgs}. For the Fixed Dropout ablation 
in Table~3 of the main paper, we use a constant dropout rate of $0.5$ 
(matching the average masking strength of LAD over the annealing 
horizon) to provide a controlled comparison.

\subsection{Hardware and Runtime}
\label{app:hardware}

All experiments are conducted on a single NVIDIA RTX 4090 GPU 
($24$ GB VRAM) with PyTorch~$2.1$ and CUDA~$12.1$. Peak GPU memory 
during training is approximately $8$--$12$ GB, depending on the 
Gaussian count.

\paragraph{Per-iteration cost.} 
Per-iteration training time depends on the number of active Gaussian 
primitives and varies across both methods and training stages. 
The dominant cost is differentiable rasterization, while LAD mask 
sampling and SAES population monitoring contribute negligibly to 
the per-iteration runtime (under $1\%$ combined).

\paragraph{End-to-end training time.} 
Total wallclock time for a single LADES training run on FIPS 25-view 
ranges from approximately $3.5$ to $4.0$ minutes, with $T_{\text{final}} = 2 t_s$ 
reaching approximately $7{,}800$ to $13{,}000$ iterations depending on 
the object (walnut, pinecone, and seashell terminate at $T_{\text{final}} 
\approx 13{,}000$, $7{,}800$, and $10{,}000$, respectively). 
Wallclock time per iteration varies across both methods and training 
stages because the cost of differentiable rasterization scales with 
the number of active Gaussian primitives, which grows during 
densification. For comparison, $\text{R}^2$-Gaussian trained to the 
full 30k-iteration budget takes approximately $10.3$ minutes on the 
same hardware.

\subsection{Random Seeds and Reproducibility}
\label{app:reproducibility}

\paragraph{Seeds.} 
All experiments use a primary random seed of $42$ for the Bernoulli mask 
sampling in LAD and VCS, the FDK initialization, and the densification 
random tie-breaking. The repeated-run variations reported as $\pm$ 
terms in Tables~1--2 are obtained by running each configuration with 
three random seeds $\{42, 52, 62\}$ and reporting the standard deviation 
across the three runs.

\paragraph{Determinism.} 
We enable PyTorch's deterministic algorithm flag where supported. 
Some CUDA operations in the differentiable rasterizer are not fully 
deterministic, leading to small ($<$$0.05$ dB) numerical fluctuations 
across repeated runs on the same seed. The $\pm$ terms reported in 
the main paper account for this fluctuation.

\paragraph{Code release.} 
An anonymized code package containing the full LADES implementation, 
the modified $\text{R}^2$-Gaussian backbone, and reproduction scripts 
for all main-paper tables and figures is provided in the supplementary 
material. The FIPS dataset is publicly available from~\cite{FIPS_CT_dataset}.

\section{Extended Experimental Results}
\label{app:extended_results}

This section provides detailed results that complement the averaged 
numbers reported in the main paper. We present (D.1) per-object 
quantitative results across all sparse-view settings, (D.2) extended 
25-view comparisons including all Gaussian-based baselines and ablation 
variants, (D.3) per-object structural diagnostics, (D.4) SAES trigger 
iterations, (D.5) training time analysis, and (D.6) correlation 
analysis between structural diagnostics and PVFD severity.

\subsection{Per-Object Quantitative Results across View Counts}
\label{app:per_object}

Tables~\ref{tab:per_object_25}--\ref{tab:per_object_10} report 
per-object 3D PSNR and 3D SSIM under the 25-, 20-, and 10-view sparse 
settings, comparing the $\text{R}^2$-Gaussian backbone with LADES. 
The averaged values in Tables~1--2 of the main paper are computed 
over the three FIPS objects (walnut, pine, seashell). The per-object 
breakdown shows that LADES outperforms the backbone on every object 
and every view count, indicating that the improvement is not driven 
by a single favorable scene.

\begin{table}[h]
\centering
\caption{Per-object 3D PSNR (dB) and 3D SSIM under the 25-view sparse 
setting on FIPS. Best results are in \textbf{bold}.}
\label{tab:per_object_25}
\small
\renewcommand{\arraystretch}{1.15}
\begin{tabular}{l cc cc cc}
\toprule
\multirow{2}{*}{\textbf{Method}} & \multicolumn{2}{c}{\textbf{Walnut}} & \multicolumn{2}{c}{\textbf{Pine}} & \multicolumn{2}{c}{\textbf{Seashell}} \\
\cmidrule(lr){2-3} \cmidrule(lr){4-5} \cmidrule(lr){6-7}
& PSNR$\uparrow$ & SSIM$\uparrow$ & PSNR$\uparrow$ & SSIM$\uparrow$ & PSNR$\uparrow$ & SSIM$\uparrow$ \\
\midrule
$\text{R}^2$-Gaussian & 28.74 & 0.663 & 38.08 & 0.925 & 39.52 & 0.937 \\
\textbf{LADES (Ours)} & \textbf{29.65} & \textbf{0.685} & \textbf{38.95} & \textbf{0.928} & \textbf{40.38} & \textbf{0.945} \\
\midrule
$\Delta$ & $+0.91$ & $+0.022$ & $+0.87$ & $+0.003$ & $+0.86$ & $+0.008$ \\
\bottomrule
\end{tabular}
\end{table}

\begin{table}[h]
\centering
\caption{Per-object 3D PSNR (dB) and 3D SSIM under the 20-view sparse 
setting on FIPS. Best results are in \textbf{bold}.}
\label{tab:per_object_20}
\small
\renewcommand{\arraystretch}{1.15}
\begin{tabular}{l cc cc cc}
\toprule
\multirow{2}{*}{\textbf{Method}} & \multicolumn{2}{c}{\textbf{Walnut}} & \multicolumn{2}{c}{\textbf{Pine}} & \multicolumn{2}{c}{\textbf{Seashell}} \\
\cmidrule(lr){2-3} \cmidrule(lr){4-5} \cmidrule(lr){6-7}
& PSNR$\uparrow$ & SSIM$\uparrow$ & PSNR$\uparrow$ & SSIM$\uparrow$ & PSNR$\uparrow$ & SSIM$\uparrow$ \\
\midrule
$\text{R}^2$-Gaussian & 27.88 & 0.650 & 36.56 & 0.916 & 37.28 & 0.928 \\
\textbf{LADES (Ours)} & \textbf{28.79} & \textbf{0.672} & \textbf{37.52} & \textbf{0.920} & \textbf{37.96} & \textbf{0.937} \\
\midrule
$\Delta$ & $+0.91$ & $+0.022$ & $+0.96$ & $+0.004$ & $+0.68$ & $+0.009$ \\
\bottomrule
\end{tabular}
\end{table}

\begin{table}[h]
\centering
\caption{Per-object 3D PSNR (dB) and 3D SSIM under the 10-view sparse 
setting on FIPS. Best results are in \textbf{bold}.}
\label{tab:per_object_10}
\small
\renewcommand{\arraystretch}{1.15}
\begin{tabular}{l cc cc cc}
\toprule
\multirow{2}{*}{\textbf{Method}} & \multicolumn{2}{c}{\textbf{Walnut}} & \multicolumn{2}{c}{\textbf{Pine}} & \multicolumn{2}{c}{\textbf{Seashell}} \\
\cmidrule(lr){2-3} \cmidrule(lr){4-5} \cmidrule(lr){6-7}
& PSNR$\uparrow$ & SSIM$\uparrow$ & PSNR$\uparrow$ & SSIM$\uparrow$ & PSNR$\uparrow$ & SSIM$\uparrow$ \\
\midrule
$\text{R}^2$-Gaussian & 25.30 & 0.612 & 33.06 & 0.885 & 33.53 & 0.902 \\
\textbf{LADES (Ours)} & \textbf{25.83} & \textbf{0.632} & \textbf{33.90} & \textbf{0.888} & \textbf{34.04} & \textbf{0.909} \\
\midrule
$\Delta$ & $+0.53$ & $+0.020$ & $+0.84$ & $+0.003$ & $+0.51$ & $+0.007$ \\
\bottomrule
\end{tabular}
\end{table}

\begin{figure}[h]
\centering
\includegraphics[width=\linewidth]{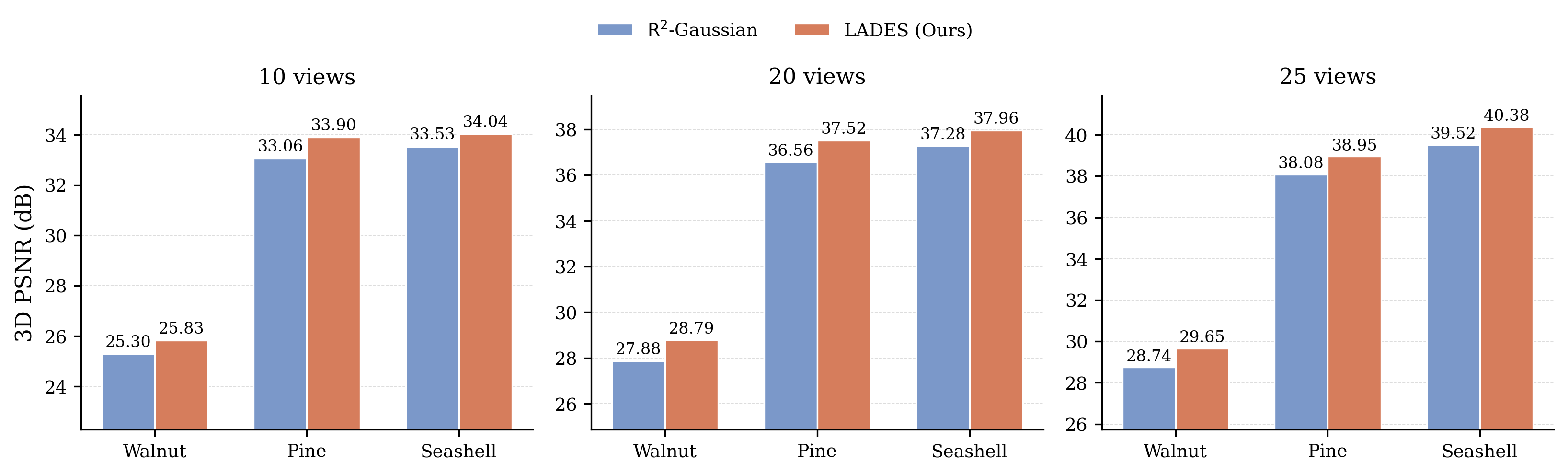}
\caption{Per-object 3D PSNR comparison between $\text{R}^2$-Gaussian 
and LADES across 10-, 20-, and 25-view sparse settings on FIPS. LADES 
consistently outperforms the backbone across all $3 \times 3 = 9$ 
combinations of object and view count.}
\label{fig:per_object_psnr}
\end{figure}

Figure~\ref{fig:per_object_psnr} visualizes the per-object PSNR results 
across all three view counts.


\paragraph{Consistency across view counts.} 
The improvement of LADES over the $\text{R}^2$-Gaussian backbone is 
consistent across all view counts and objects, with average 3D PSNR 
gains of $+0.63$ dB at 10 views, $+0.85$ dB at 20 views, and $+0.88$ dB 
at 25 views. The relative improvement is somewhat smaller under the 
most extreme sparsity (10 views), where all methods are constrained 
by the larger projection-domain null space discussed in 
Appendix~\ref{app:formal_analysis}. Even in this challenging regime, 
LADES maintains a positive gain on every object.

\paragraph{Difficulty across objects.} 
Walnut consistently yields the lowest 3D PSNR across all methods 
(approximately $25$--$30$ dB), reflecting its high internal contrast 
between shell and kernel and the resulting boundary complexity. 
Pine and Seashell yield substantially higher 3D PSNR (approximately 
$33$--$40$ dB), with Seashell slightly higher at 25 views. The 
relative gain of LADES is comparable across all three objects, 
indicating that PVFD-aware training control benefits the reconstruction 
of both high-contrast and lower-contrast structures.

\subsection{Per-Object Breakdown of 25-View Comparison}
\label{app:per_object_25_full}

Table~\ref{tab:per_object_25_full} provides the per-object PSNR 
breakdown for the full 25-view comparison reported in Table~1 of 
the main paper. This complements the averaged comparison and shows 
that LADES achieves the best PSNR on every object across all evaluated 
methods.

\begin{table}[h]
\centering
\caption{Per-object 3D PSNR (dB) under the 25-view sparse setting on 
FIPS, including all Gaussian-based baselines and ablation variants 
evaluated in the main paper. Best results are in \textbf{bold}.}
\label{tab:per_object_25_full}
\small
\renewcommand{\arraystretch}{1.15}
\begin{tabular}{l ccc c}
\toprule
\textbf{Method} & \textbf{Walnut} & \textbf{Pine} & \textbf{Seashell} & \textbf{Avg} \\
\midrule
$\text{R}^2$-Gaussian   & 28.74 & 38.08 & 39.52 & 35.45 \\
TAG-Gaussian            & 28.93 & 38.32 & 39.80 & 35.68 \\
Fixed Dropout           & 26.05 & 32.73 & 33.09 & 30.62 \\
LAD-only                & 29.54 & 38.77 & 39.71 & 36.01 \\
SAES-only               & 29.46 & 38.66 & 40.25 & 36.12 \\
\textbf{LADES (Ours)}   & \textbf{29.65} & \textbf{38.95} & \textbf{40.38} & \textbf{36.33} \\
\bottomrule
\end{tabular}
\end{table}

\paragraph{Observations.} 
Three patterns emerge from the per-object breakdown. First, LADES 
achieves the best PSNR on every object, confirming that the averaged 
gain in Table~1 of the main paper is not driven by a single favorable 
scene. Second, Fixed Dropout severely underfits all three objects 
(losing $2.7$--$6.4$ dB compared to the backbone), confirming that 
constant masking restricts representational capacity throughout 
training. Third, both LAD-only and SAES-only individually outperform 
the backbone on every object, and combining them in LADES yields 
further gains, indicating that the two components address 
complementary failure modes.

\subsection{Per-Object Structural Diagnostics}
\label{app:per_object_diagnostics}

Tables~\ref{tab:per_object_diag} and~\ref{tab:per_object_vcs} report 
per-object structural diagnostics under the 25-view setting. For 
LADES, the training trajectory is monotonically non-decreasing in 
3D PSNR up to $T_{\text{final}}$, and we report $\Delta_{\text{PVFD}}$ 
as zero (no observed post-peak degradation along the SAES-controlled 
trajectory).

\begin{table}[h]
\centering
\caption{Per-object PVFD severity and primitive-level diagnostics 
under the 25-view setting on FIPS. $t^{\star}$ denotes the volumetric 
peak iteration, $\Delta_{\text{PVFD}}$ the post-peak 3D PSNR degradation, 
$\overline{\text{GAI}}$ the field-level mean Geometric Anisotropy 
Index, and $R_{\text{needle}}$ the fraction of needle-like primitives 
($\text{GAI}_i > 50$).}
\label{tab:per_object_diag}
\small
\renewcommand{\arraystretch}{1.15}
\begin{tabular}{l c rrrr}
\toprule
\textbf{Method} & \textbf{Object} & $t^{\star}$ & $\Delta_{\text{PVFD}}\downarrow$ & $\overline{\text{GAI}}\downarrow$ & $R_{\text{needle}}$ (\%)$\downarrow$ \\
\midrule
\multirow{3}{*}{$\text{R}^2$-GS}
 & Walnut   & 2{,}000  & 0.705 & 14.82 & 6.26 \\
 & Pine     & 20{,}000 & 0.072 & 13.61 & 5.53 \\
 & Seashell & 6{,}000  & 0.674 & 22.84 & 11.43 \\
\midrule
\multirow{3}{*}{\textbf{LADES}}
 & Walnut   & 13{,}000 & \textbf{0.000} & \textbf{8.07} & \textbf{1.73} \\
 & Pine     & 7{,}800  & \textbf{0.000} & \textbf{4.52} & \textbf{0.07} \\
 & Seashell & 10{,}000 & \textbf{0.000} & \textbf{6.98} & \textbf{0.50} \\
\bottomrule
\end{tabular}
\end{table}

\begin{table}[h]
\centering
\caption{Per-object Volumetric Co-adaptation Score (VCS) under the 
25-view setting on FIPS. Lower is better.}
\label{tab:per_object_vcs}
\small
\renewcommand{\arraystretch}{1.15}
\begin{tabular}{l ccc}
\toprule
\textbf{Method} & \textbf{Walnut} & \textbf{Pine} & \textbf{Seashell} \\
\midrule
$\text{R}^2$-GS & $6.57 \times 10^{-4}$ & $2.82 \times 10^{-4}$ & $1.78 \times 10^{-4}$ \\
\textbf{LADES} & $\mathbf{1.12 \times 10^{-4}}$ & $\mathbf{5.16 \times 10^{-5}}$ & $\mathbf{3.24 \times 10^{-5}}$ \\
\bottomrule
\end{tabular}
\end{table}

\paragraph{Observations.} 
Three patterns emerge from the per-object diagnostics. First, the 
volumetric peak iteration $t^{\star}$ of $\text{R}^2$-GS varies 
substantially across objects ($2{,}000$ for Walnut, $6{,}000$ for 
Seashell, and $20{,}000$ for Pine), with $\Delta_{\text{PVFD}}$ 
correspondingly ranging from $0.072$ to $0.705$ dB. This indicates 
that the timing and severity of PVFD are object-dependent, and that 
a fixed-iteration training budget is poorly suited to handle this 
variation. Second, $\overline{\text{GAI}}$ and $R_{\text{needle}}$ 
are reduced by factors of approximately $2$--$3\times$ and $5$--$80\times$ 
respectively under LADES, with the largest relative reduction in 
$R_{\text{needle}}$ on Pine (from $5.53\%$ to $0.07\%$). Third, VCS 
is reduced by approximately $5$--$6\times$ on every object, indicating 
that LADES produces volumetric representations that are uniformly 
more robust under primitive perturbation.

\subsection{SAES Trigger Iterations}
\label{app:saes_iterations}

Table~\ref{tab:saes_trigger} reports the SAES trigger iteration $t_s$ 
and the final training iteration $T_{\text{final}} = 2 t_s$ for each 
object and view count under LADES. SAES is detected from the 
sliding-window-averaged Gaussian population growth rate falling below 
the threshold $\tau = 10^{-3}$, without using any ground-truth 
volumetric metric.

\begin{table}[h]
\centering
\caption{SAES trigger iteration $t_s$ and final training iteration 
$T_{\text{final}} = 2 t_s$ under LADES, across objects and view counts.}
\label{tab:saes_trigger}
\small
\renewcommand{\arraystretch}{1.15}
\begin{tabular}{l cc cc cc}
\toprule
\multirow{2}{*}{\textbf{Object}} & \multicolumn{2}{c}{\textbf{10 views}} & \multicolumn{2}{c}{\textbf{20 views}} & \multicolumn{2}{c}{\textbf{25 views}} \\
\cmidrule(lr){2-3} \cmidrule(lr){4-5} \cmidrule(lr){6-7}
& $t_s$ & $T_{\text{final}}$ & $t_s$ & $T_{\text{final}}$ & $t_s$ & $T_{\text{final}}$ \\
\midrule
Walnut   & 6{,}300 & 12{,}600 & 6{,}500 & 13{,}000 & 6{,}500 & 13{,}000 \\
Pine     & 3{,}600 & 7{,}200  & 4{,}000 & 8{,}000  & 3{,}900 & 7{,}800  \\
Seashell & 4{,}600 & 9{,}200  & 4{,}600 & 9{,}200  & 5{,}000 & 10{,}000 \\
\midrule
\textbf{Mean} & \textbf{4{,}833} & \textbf{9{,}667} & \textbf{5{,}033} & \textbf{10{,}067} & \textbf{5{,}133} & \textbf{10{,}267} \\
\bottomrule
\end{tabular}
\end{table}

\paragraph{Observations.} 
The SAES trigger iteration is remarkably stable across view counts 
for the same object: Walnut consistently triggers around 
$t_s \approx 6{,}300$--$6{,}500$, Pine around $t_s \approx 3{,}600$--$4{,}000$, 
and Seashell around $t_s \approx 4{,}600$--$5{,}000$. This stability 
across view counts indicates that SAES detects an intrinsic, 
object-dependent signal of structural saturation that is largely 
decoupled from the specific projection sampling density. In contrast, 
the trigger iteration varies more across objects, with structurally 
simpler scenes (Pine) saturating earlier than scenes with complex 
boundaries (Walnut). All trigger iterations are well below the 
$30{,}000$-iteration training budget used by baselines.

\subsection{Training Time Analysis}
\label{app:training_time}

Table~\ref{tab:time_iter} compares the total wallclock training time 
and final iteration count for all 3DGS-based methods under the 25-view 
sparse setting.

\begin{table}[h]
\centering
\caption{Training time and final iteration count under the 25-view 
sparse setting on FIPS, averaged across three objects.}
\label{tab:time_iter}
\small
\renewcommand{\arraystretch}{1.15}
\begin{tabular}{l cc}
\toprule
\textbf{Method} & \textbf{Training time (min)} & \textbf{Final iteration} \\
\midrule
$\text{R}^2$-Gaussian & 10.3 & 30{,}000 \\
TAG-Gaussian          & 10.2 & 30{,}000 \\
Dropout-GS            & 19.5 & 30{,}000 \\
GR-Gaussian           & 11.4 & 30{,}000 \\
\textbf{LADES (Ours)} & \textbf{3.7} & $\sim$$10{,}300$ \\
\bottomrule
\end{tabular}
\end{table}

\begin{figure}[h]
\centering
\includegraphics[width=0.7\linewidth]{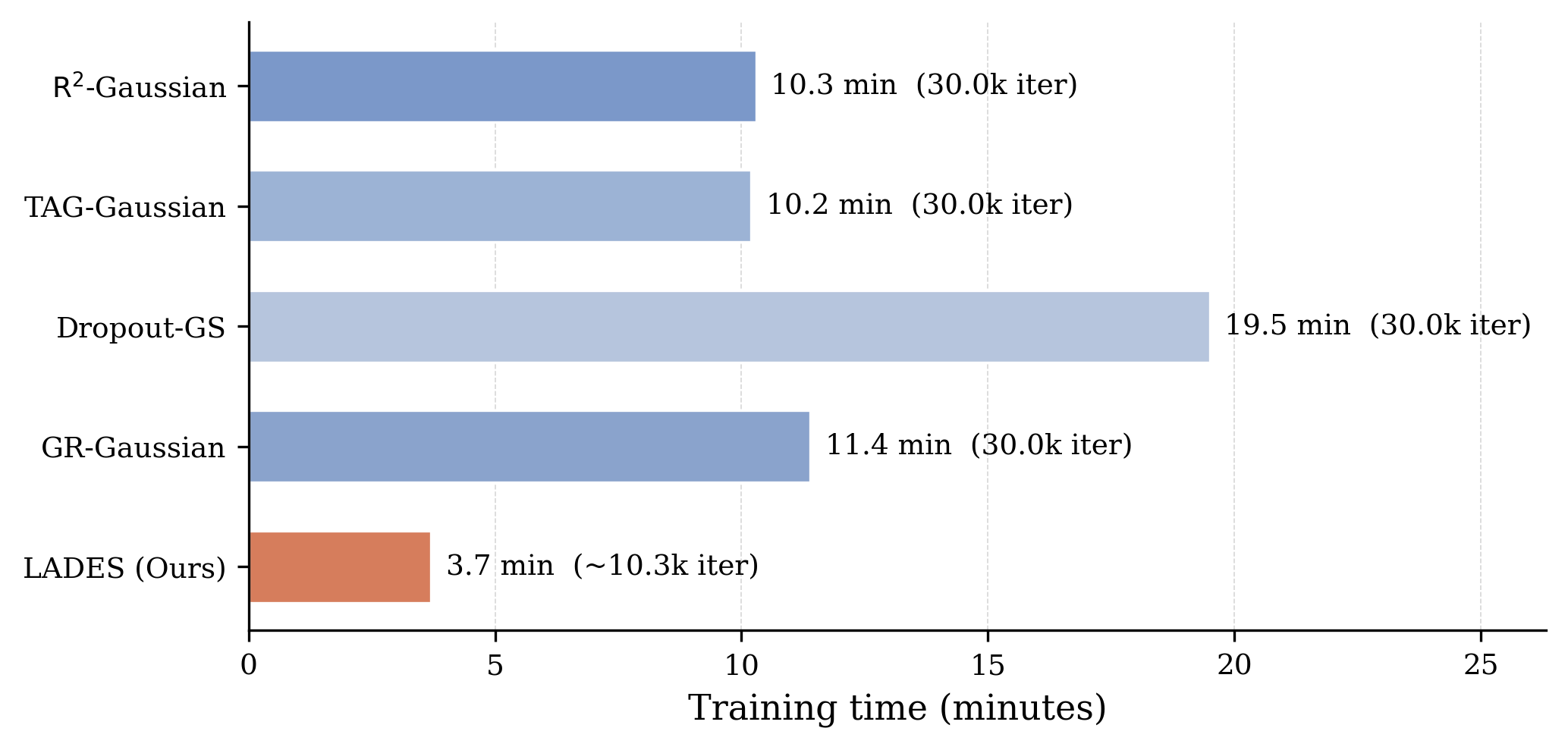}
\caption{Training time comparison under the 25-view sparse setting 
on FIPS, averaged across three objects. LADES achieves a $\sim$$64\%$ 
reduction in wallclock time compared to the $\text{R}^2$-Gaussian 
backbone, while simultaneously achieving higher 3D PSNR. The reduction 
comes from SAES detecting structural saturation and stopping training 
at $T_{\text{final}} = 2 t_s \approx 10{,}300$ iterations rather than 
running to the fixed 30k-iteration budget.}
\label{fig:training_time}
\end{figure}

\paragraph{On per-iteration cost.} 
We note that per-iteration wallclock time varies across both methods 
and training stages, because the cost of differentiable rasterization 
scales with the number of active Gaussian primitives, which itself 
grows during densification. Consequently, the wallclock comparison 
in Table~\ref{tab:time_iter} reflects total end-to-end runtime on 
identical hardware rather than a simple multiplication of iteration 
count and a fixed per-iteration cost.

\paragraph{Source of efficiency improvement.} 
LADES reduces total training time by approximately $64\%$ compared 
to $\text{R}^2$-Gaussian under the 25-view setting, while simultaneously 
achieving higher 3D PSNR. The reduction comes from SAES detecting 
structural saturation and stopping densification at $t_s \approx 5{,}000$, 
followed by fixed-topology refinement until $T_{\text{final}} = 2 t_s 
\approx 10{,}000$. This avoids the redundant late-stage densification 
of the baseline, which continues to add and refine primitives until 
the fixed $30{,}000$-iteration budget is exhausted---adding training 
cost without improving (and in fact harming) volumetric fidelity, 
as quantified by $\Delta_{\text{PVFD}}$ in 
Table~\ref{tab:per_object_diag}.

\subsection{Correlation between Structural Diagnostics and PVFD Severity}
\label{app:diagnostic_correlation}

\begin{figure}[h]
\centering
\includegraphics[width=\linewidth]{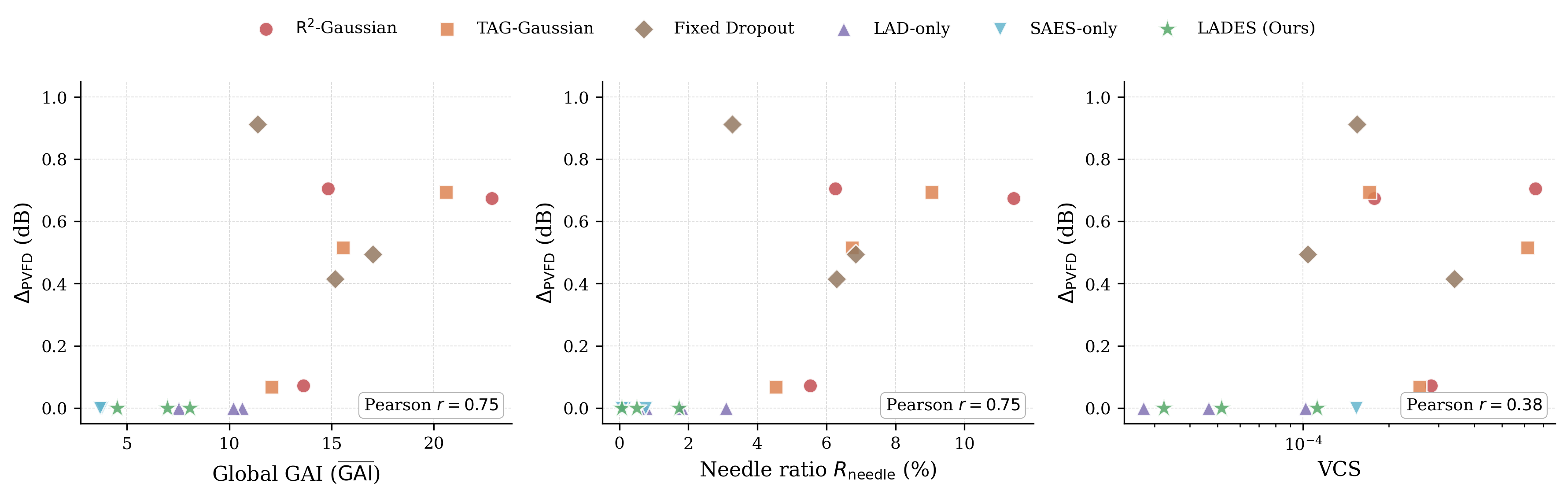}
\caption{Correlation between structural diagnostics and PVFD severity 
$\Delta_{\text{PVFD}}$ across all (method, object) configurations under 
the 25-view setting. Each point represents one configuration. LADES 
points cluster near the origin (low diagnostic values, no observable 
PVFD), while baseline configurations spread toward higher diagnostic 
values and larger $\Delta_{\text{PVFD}}$. Pearson correlation 
coefficients are computed across all 18 configurations.}
\label{fig:diag_correlation}
\end{figure}

A natural validation of the structural diagnostics 
($\overline{\text{GAI}}$, $R_{\text{needle}}$, VCS) is to test whether 
they correlate with the observed PVFD severity $\Delta_{\text{PVFD}}$. 
Figure~\ref{fig:diag_correlation} plots each diagnostic against 
$\Delta_{\text{PVFD}}$ across all (method, object) combinations 
under the 25-view setting.

The scatter plot shows a positive trend in all three panels: methods 
with higher $\overline{\text{GAI}}$, $R_{\text{needle}}$, and VCS 
also exhibit larger $\Delta_{\text{PVFD}}$. LADES configurations 
cluster near the origin (low diagnostic values, no observable PVFD), 
while $\text{R}^2$-Gaussian and Fixed Dropout configurations spread 
toward the upper right. This pattern suggests that the proposed 
diagnostics capture meaningful structural correlates of PVFD rather 
than ad-hoc descriptors of the Gaussian field, supporting their use 
as ground-truth-free indicators of training health.

\paragraph{Quantitative correlation.} 
Across all 18 (method, object) configurations, $\overline{\text{GAI}}$ 
and $R_{\text{needle}}$ both exhibit strong positive correlation with 
$\Delta_{\text{PVFD}}$ (Pearson $r = 0.75$ for both), supporting their 
use as primitive-level indicators of PVFD risk. The correlation between 
VCS and $\Delta_{\text{PVFD}}$ is more moderate ($r = 0.38$), reflecting 
that volume-level fragility captures a complementary, partially 
orthogonal aspect of structural degeneration: a representation can be 
locally anisotropic without being globally fragile, and vice versa. 
We recommend reporting all three diagnostics jointly when monitoring 
sparse-view 3DGS-CT training.

\section{Diagnostic Metrics: Validation and Sensitivity}
\label{app:diagnostic_validation}

We introduced two structural diagnostics in Section~3.2 of the main 
paper: the Geometric Anisotropy Index (GAI) and its derivatives 
($\overline{\text{GAI}}$, $\text{GAI}_{\max}$, $R_{\text{needle}}$), 
and the Volumetric Co-adaptation Score (VCS). Beyond their qualitative 
observation in the main paper, this section provides a more thorough 
validation: (E.1) implementation details of the diagnostics, 
(E.2) sensitivity to the GAI needle threshold $\tau_{\text{GAI}}$, 
(E.3) sensitivity to the VCS sampling parameters $p_{\text{vcs}}$ 
and $K_{\text{vcs}}$, (E.4) joint information across the three 
diagnostics, and (E.5) discussion of alternative diagnostic designs 
we considered.

\subsection{Implementation Details of the Diagnostics}
\label{app:diagnostic_impl}

For completeness, we describe the implementation of the diagnostics 
used throughout this paper.

\paragraph{GAI implementation.} 
For each Gaussian primitive $i$, we read the post-activation principal 
scales $(s_{i,x}, s_{i,y}, s_{i,z})$ directly from the optimized 
3DGS parameters. The per-primitive GAI is computed as in Eq.~(3) of 
the main paper with $\epsilon = 10^{-8}$. $\overline{\text{GAI}}$, 
$\text{GAI}_{\max}$, and $R_{\text{needle}}$ are computed by averaging, 
taking the maximum, and thresholding ($\tau_{\text{GAI}} = 50$) over 
all active primitives.

\paragraph{VCS implementation.} 
We adopt a density-suppression formulation rather than a hard mask. 
At each Monte Carlo trial $k$, for primitives selected by the 
Bernoulli mask ($m_i^{(k)} = 0$), we set the pre-activation density 
parameter to a large negative value such that the post-Softplus 
density is numerically zero, and re-query the voxelized volume 
$V^{(k)}(\mathbf{x})$. This produces the same effect as 
``removing'' the primitive from the rasterization while preserving 
all other parameters and the rendering path. After each trial, the 
original density values are restored before the next trial.

\paragraph{Ground-truth-free foreground.} 
The foreground region $\Omega_{\text{fg}}$ used to average the 
voxel-wise variance is determined directly from the reconstructed 
volume rather than from the ground truth. Specifically, we compute 
the mean voxel density 
$\bar{V}(\mathbf{x}) = \frac{1}{K_{\text{vcs}}} \sum_k V^{(k)}(\mathbf{x})$ 
across the $K_{\text{vcs}}$ trials, and define 
$\Omega_{\text{fg}} = \{ \mathbf{x} : \bar{V}(\mathbf{x}) > \tau_{\text{fg}} \}$ 
with $\tau_{\text{fg}} = 0.01$ in the normalized $[0,1]$ density 
range. VCS is therefore computed entirely from the optimized 
Gaussian field, without access to the ground-truth volume, and is 
consistent with the ground-truth-free design philosophy of LADES.

\paragraph{Numerical evaluation.} 
The integral in Eq.~(6) of the main paper is evaluated as a discrete 
sum over voxels in $\Omega_{\text{fg}}$, normalized by the foreground 
voxel count $|\Omega_{\text{fg}}|$.

\subsection{Sensitivity to the GAI Needle Threshold }
\label{app:gai_threshold}

The needle ratio $R_{\text{needle}}$ requires choosing a threshold 
$\tau_{\text{GAI}}$ above which a primitive is classified as 
needle-like. In the main paper we use $\tau_{\text{GAI}} = 50$. 
Table~\ref{tab:gai_threshold} reports $R_{\text{needle}}$ under 
$\tau_{\text{GAI}} \in \{20, 30, 50, 80, 100\}$ for $\text{R}^2$-GS 
and LADES under the 25-view setting (averaged across three objects).

\begin{table}[h]
\centering
\caption{Sensitivity of $R_{\text{needle}}$ (in \%) to the threshold 
$\tau_{\text{GAI}}$ under the 25-view setting on FIPS, averaged 
across three objects. The relative ordering between methods is 
preserved across all threshold choices, with LADES yielding 
substantial reductions at every threshold.}
\label{tab:gai_threshold}
\small
\renewcommand{\arraystretch}{1.15}
\begin{tabular}{l ccccc}
\toprule
\textbf{Method} & $\tau_{\text{GAI}} = 20$ & $\tau_{\text{GAI}} = 30$ 
                & $\boldsymbol{\tau_{\text{GAI}} = 50}$ 
                & $\tau_{\text{GAI}} = 80$ & $\tau_{\text{GAI}} = 100$ \\
\midrule
$\text{R}^2$-GS & 17.12 & 12.24 & \textbf{7.74} & 4.60 & 3.48 \\
LADES           &  4.49 &  2.04 & \textbf{0.77} & 0.28 & 0.15 \\
\midrule
Reduction (×) & $3.8\times$ & $6.0\times$ & $\mathbf{10.1\times}$ 
              & $16.6\times$ & $23.4\times$ \\
\bottomrule
\end{tabular}
\end{table}

\paragraph{Observation.} 
The relative ordering between $\text{R}^2$-GS and LADES is preserved 
across all values of $\tau_{\text{GAI}}$. The reduction factor 
\emph{increases} monotonically with the threshold, ranging from 
$3.8\times$ at $\tau_{\text{GAI}} = 20$ to $23.4\times$ at 
$\tau_{\text{GAI}} = 100$. This indicates that LADES suppresses 
extreme primitive elongation more aggressively than mild 
anisotropy: the most pathological needle-like primitives 
($\text{GAI} > 100$) are reduced by more than an order of magnitude. 
The default choice of $\tau_{\text{GAI}} = 50$ is therefore not a 
narrowly tuned setting, and the reported PVFD diagnosis would hold 
qualitatively under any reasonable threshold in this range.

\subsection{Sensitivity to VCS Sampling Parameters}
\label{app:vcs_sensitivity}

VCS depends on two sampling hyperparameters: the Bernoulli drop 
probability $p_{\text{vcs}}$ and the number of Monte Carlo trials 
$K_{\text{vcs}}$. The default values used in the main paper are 
$p_{\text{vcs}} = 0.2$ and $K_{\text{vcs}} = 10$.

\paragraph{Sensitivity to $p_{\text{vcs}}$.} 
Table~\ref{tab:vcs_pvcs} reports VCS for 
$p_{\text{vcs}} \in \{0.05, 0.10, 0.20, 0.30, 0.50\}$ on 
$\text{R}^2$-GS and LADES under the 25-view setting 
($K_{\text{vcs}} = 10$).

\begin{table}[h]
\centering
\caption{Sensitivity of VCS to the drop probability $p_{\text{vcs}}$ 
under the 25-view setting on FIPS, averaged across three objects. 
Values are reported in units of $10^{-4}$.}
\label{tab:vcs_pvcs}
\small
\renewcommand{\arraystretch}{1.15}
\begin{tabular}{l ccccc}
\toprule
\textbf{Method} & $p = 0.05$ & $p = 0.10$ & $\boldsymbol{p = 0.20}$ 
                & $p = 0.30$ & $p = 0.50$ \\
\midrule
$\text{R}^2$-GS ($\times 10^{-4}$) 
                & 0.94 & 1.84 & \textbf{3.35} & 4.56 & 5.82 \\
LADES ($\times 10^{-4}$) 
                & 0.17 & 0.32 & \textbf{0.59} & 0.79 & 1.02 \\
\midrule
Reduction (×)   & $5.6\times$ & $5.8\times$ & $\mathbf{5.7\times}$ 
                & $5.8\times$ & $5.7\times$ \\
\bottomrule
\end{tabular}
\end{table}

\paragraph{Bernoulli scaling sanity check.} 
For a Bernoulli-perturbation diagnostic, the variance contribution 
of each independently dropped primitive is proportional to 
$p_{\text{vcs}}(1 - p_{\text{vcs}})$ in the limit of independent 
primitives. We compare the empirical scaling with this prediction 
in Table~\ref{tab:vcs_scaling}.

\begin{table}[h]
\centering
\caption{VCS values normalized by the value at $p_{\text{vcs}} = 0.2$, 
compared with the theoretical Bernoulli-variance prediction $p(1-p)$ 
(also normalized at $p = 0.2$). Both methods follow the predicted 
scaling closely, with a slight excess at $p \ge 0.3$ attributable 
to the nonlinear voxelization of the suppressed-density formulation.}
\label{tab:vcs_scaling}
\small
\renewcommand{\arraystretch}{1.15}
\begin{tabular}{l ccccc}
\toprule
& $p = 0.05$ & $p = 0.10$ & $p = 0.20$ & $p = 0.30$ & $p = 0.50$ \\
\midrule
$\text{R}^2$-GS & 0.280 & 0.549 & 1.000 & 1.360 & 1.738 \\
LADES           & 0.287 & 0.543 & 1.000 & 1.346 & 1.739 \\
\midrule
$p(1-p)$ (theory) & 0.297 & 0.562 & 1.000 & 1.312 & 1.562 \\
\bottomrule
\end{tabular}
\end{table}

The empirical scaling is in close agreement with the theoretical 
$p(1-p)$ prediction in the small-to-moderate regime ($p \le 0.2$), 
and deviates only mildly at larger $p$. Importantly, the relative 
reduction of LADES over $\text{R}^2$-GS is essentially constant at 
$\sim$$5.7\times$ across the entire $p_{\text{vcs}}$ range, indicating 
that the VCS improvement of LADES reflects an intrinsic robustness 
of the optimized representation rather than an artifact of a 
particular sampling strength. This consistency, together with the 
agreement with Bernoulli-variance theory, supports VCS as a 
principled rather than ad-hoc diagnostic.

\paragraph{Sensitivity to $K_{\text{vcs}}$.} 
Table~\ref{tab:vcs_kvcs} reports VCS for 
$K_{\text{vcs}} \in \{5, 10, 20, 50\}$ to assess the Monte Carlo 
estimation variance ($p_{\text{vcs}} = 0.2$).

\begin{table}[h]
\centering
\caption{Sensitivity of VCS to the number of Monte Carlo trials 
$K_{\text{vcs}}$ under the 25-view setting ($p_{\text{vcs}} = 0.2$), 
averaged across three objects. Values are in units of $10^{-4}$. 
Relative changes are reported with respect to $K = 10$.}
\label{tab:vcs_kvcs}
\small
\renewcommand{\arraystretch}{1.15}
\begin{tabular}{l cccc}
\toprule
\textbf{Method} & $K = 5$ & $\boldsymbol{K = 10}$ & $K = 20$ & $K = 50$ \\
\midrule
$\text{R}^2$-GS ($\times 10^{-4}$) 
                & 3.02 & \textbf{3.38} & 3.54 & 3.65 \\
LADES ($\times 10^{-4}$) 
                & 0.52 & \textbf{0.58} & 0.62 & 0.64 \\
\midrule
Relative change vs $K = 10$ ($\text{R}^2$-GS) 
                & $-10.7\%$ & --- & $+4.7\%$ & $+8.0\%$ \\
Relative change vs $K = 10$ (LADES) 
                & $-11.0\%$ & --- & $+5.9\%$ & $+9.4\%$ \\
\bottomrule
\end{tabular}
\end{table}

\paragraph{Observations on $K_{\text{vcs}}$.} 
VCS estimates exhibit a small upward drift as $K_{\text{vcs}}$ 
increases, reflecting the small-sample bias of the variance estimator. 
With $K_{\text{vcs}} = 5$, VCS is underestimated by approximately 
$11\%$ on both methods relative to $K_{\text{vcs}} = 10$; with 
$K_{\text{vcs}} = 50$, VCS is overestimated by approximately $8$--$9\%$. 
The default $K_{\text{vcs}} = 10$ is a practical compromise: it 
mitigates the small-$K$ bias to a stable level while keeping the 
computational cost low enough for routine use during training 
diagnostics. Critically, the gap between $\text{R}^2$-GS and LADES 
($\sim$$5.8\times$ reduction at $K = 10$) is much larger than the 
$K_{\text{vcs}}$-induced variation in either method, so the 
qualitative comparison reported in the main paper is robust to 
this hyperparameter choice.

\subsection{Joint Information Across the Three Diagnostics}
\label{app:joint_information}

The three diagnostics $\overline{\text{GAI}}$, $R_{\text{needle}}$, 
and VCS are not redundant: they capture different aspects of 
structural degeneration. We note two complementary observations.

\paragraph{Primitive-level vs volume-level signal.} 
$\overline{\text{GAI}}$ and $R_{\text{needle}}$ both summarize 
primitive-level anisotropy and are highly correlated by construction: 
$R_{\text{needle}}$ is a thresholded counterpart of GAI. 
VCS captures volume-level fragility, which depends on \emph{how} 
primitives interact rather than how individual primitives are 
shaped. A representation can be locally anisotropic without being 
globally fragile (e.g., when needle-like primitives are well isolated), 
or globally fragile without strong individual anisotropy (e.g., 
when many low-anisotropy primitives co-adapt to fit specific 
projection residuals). Reporting both classes therefore provides a 
more complete picture of structural degeneration than relying on 
either alone.

\paragraph{Joint reduction under LADES.} 
Across all three diagnostics, LADES yields substantial reductions 
relative to the $\text{R}^2$-Gaussian backbone: on average a 
$\sim$$10\times$ reduction in $R_{\text{needle}}$ at the default 
threshold, a $\sim$$5.7\times$ reduction in VCS at the default 
sampling, and a $\sim$$2.6\times$ reduction in $\overline{\text{GAI}}$ 
(Table~2 of the main paper). The fact that all three indicators 
move consistently---rather than only one of them---supports the 
interpretation that LADES improves a unified notion of structural 
health rather than merely optimizing one specific diagnostic.

\subsection{Alternative Diagnostic Designs}
\label{app:alternative_diagnostics}

We briefly discuss several alternative diagnostic designs that we 
considered but did not adopt as primary metrics.

\paragraph{Covariance condition number as an alternative to GAI.} 
Instead of the max-to-min scale ratio, one could use the condition 
number $\kappa(\Sigma_i) = \lambda_{\max}/\lambda_{\min}$ of the full 
covariance matrix $\Sigma_i$ of each Gaussian primitive. For 
diagonal-rotated 3DGS primitives, this is mathematically equivalent 
to $(s_{\max}/s_{\min})^2$, i.e., the squared GAI. The squared 
formulation accentuates extreme anisotropy more strongly but is 
less stable numerically near the cutoff. We adopted the linear 
formulation because (i) it is more interpretable as a length-scale 
ratio and (ii) it produces a more uniform distribution of values 
useful for thresholding.

\paragraph{Density entropy as an alternative to VCS.} 
A natural alternative to VCS is the entropy of the voxelized density 
field, $H(V) = -\int \tilde{V}(\mathbf{x}) \log \tilde{V}(\mathbf{x}) 
\,d\mathbf{x}$ where $\tilde{V}$ is the normalized density. We found 
that density entropy is dominated by the bulk content of the volume 
and is relatively insensitive to the small subset of fragile primitives 
that drive PVFD. VCS, in contrast, directly probes the dependence of 
the density field on individual primitives via Bernoulli perturbation, 
and is therefore more diagnostically sharp.

\paragraph{Gradient-based fragility as an alternative to VCS.} 
A more efficient surrogate for VCS would be the per-voxel sensitivity 
$\|\partial V / \partial \theta_i\|$ where $\theta_i$ are the parameters 
of primitive $i$. This avoids Monte Carlo sampling but requires storing 
and aggregating per-primitive Jacobians, which is memory-intensive at 
scale. We leave gradient-based diagnostics as a direction for future 
work and adopt the Bernoulli-sampling formulation for its simplicity 
and direct interpretability as a robustness measure.

\subsection{Summary}
\label{app:diagnostic_summary}

The validation results in this section support three claims about 
the proposed diagnostics. First, both $R_{\text{needle}}$ and VCS 
show substantial separation between $\text{R}^2$-GS and LADES under 
all reasonable hyperparameter settings, indicating that the diagnostic 
contrast is not an artifact of narrowly tuned thresholds 
(Tables~\ref{tab:gai_threshold}--\ref{tab:vcs_kvcs}). Second, the 
empirical scaling of VCS with $p_{\text{vcs}}$ closely matches the 
theoretical Bernoulli-variance prediction $p(1-p)$ on both methods, 
supporting VCS as a principled rather than ad-hoc indicator 
(Table~\ref{tab:vcs_scaling}). Third, primitive-level 
($R_{\text{needle}}$) and volume-level (VCS) diagnostics provide 
complementary information about structural degeneration, justifying 
their joint use rather than reliance on any single metric.


\newpage
\input{checklist.tex}

\end{document}

%% file: checklist.tex
\section*{NeurIPS Paper Checklist}

\begin{enumerate}

\item {\bf Claims}
    \item[] Question: Do the main claims made in the abstract and introduction accurately reflect the paper's contributions and scope?
    \item[] Answer: \answerYes{}
    \item[] Justification: The abstract and introduction state the paper's main claims: identifying PVFD, introducing structural diagnostics, and proposing LADES for sparse-view 3DGS-CT. The experimental results evaluate these claims on FIPS sparse-view CT reconstruction.

\item {\bf Limitations}
    \item[] Question: Does the paper discuss the limitations of the work performed by the authors?
    \item[] Answer: \answerYes{}
    \item[] Justification: The paper discusses the scope of the evaluation, including its focus on sparse-view CT, the use of FIPS data, and the need for further validation under other scanner geometries, noise levels, and clinical protocols.

\item {\bf Theory assumptions and proofs}
    \item[] Question: For each theoretical result, does the paper provide the full set of assumptions and a complete (and correct) proof?
    \item[] Answer: \answerNA{}
    \item[] Justification: The paper does not present formal theoretical theorems or proofs. The mathematical expressions define diagnostic metrics and training criteria used in the proposed empirical method.

\item {\bf Experimental result reproducibility}
    \item[] Question: Does the paper fully disclose all the information needed to reproduce the main experimental results of the paper to the extent that it affects the main claims and/or conclusions of the paper (regardless of whether the code and data are provided or not)?
    \item[] Answer: \answerYes{}
    \item[] Justification: The paper specifies the dataset, sparse-view protocols, baselines, evaluation metrics, training budget, key hyperparameters, and hardware used for the main experiments.

\item {\bf Open access to data and code}
    \item[] Question: Does the paper provide open access to the data and code, with sufficient instructions to faithfully reproduce the main experimental results, as described in supplemental material?
    \item[] Answer: \answerYes{}
    \item[] Justification: An anonymized code package and reproduction instructions are provided in the supplementary material. The FIPS dataset is publicly available and cited.

\item {\bf Experimental setting/details}
    \item[] Question: Does the paper specify all the training and test details (e.g., data splits, hyperparameters, how they were chosen, type of optimizer) necessary to understand the results?
    \item[] Answer: \answerYes{}
    \item[] Justification: The experimental setup describes the FIPS objects, sparse-view sampling settings, reconstruction resolution, baselines, metrics, GPU hardware, training budget, and LADES hyperparameters such as \(p_0\), \(T_{\mathrm{anneal}}\), \(t_{\min}\), \(\Delta\), \(W\), and \(\tau\).

\item {\bf Experiment statistical significance}
    \item[] Question: Does the paper report error bars suitably and correctly defined or other appropriate information about the statistical significance of the experiments?
    \item[] Answer: \answerYes{}
    \item[] Justification: The main quantitative results report repeated-run variation using the \(\pm\) term where available. These variations are intended to reflect repeated runs under the same experimental protocol rather than cross-object difficulty.

\item {\bf Experiments compute resources}
    \item[] Question: For each experiment, does the paper provide sufficient information on the computer resources (type of compute workers, memory, time of execution) needed to reproduce the experiments?
    \item[] Answer: \answerYes{}
    \item[] Justification: The paper reports that 3DGS-based experiments are implemented in PyTorch and trained on a single NVIDIA RTX 4090 GPU. Runtime is reported where available in the experimental tables.

\item {\bf Code of ethics}
    \item[] Question: Does the research conducted in the paper conform, in every respect, with the NeurIPS Code of Ethics \url{https://neurips.cc/public/EthicsGuidelines}?
    \item[] Answer: \answerYes{}
    \item[] Justification: The work uses public CT object data and does not involve private data, human-subject studies, unsafe deployment, or high-risk model release. The paper does not make clinical deployment claims.

\item {\bf Broader impacts}
    \item[] Question: Does the paper discuss both potential positive societal impacts and negative societal impacts of the work performed?
    \item[] Answer: \answerYes{}
    \item[] Justification: The work may benefit sparse-view tomographic reconstruction by improving volumetric fidelity. A potential negative impact is that unvalidated reconstruction methods could be harmful if used in safety-critical or clinical settings; the paper explicitly avoids claiming clinical deployment readiness.

\item {\bf Safeguards}
    \item[] Question: Does the paper describe safeguards that have been put in place for responsible release of data or models that have a high risk for misuse (e.g., pre-trained language models, image generators, or scraped datasets)?
    \item[] Answer: \answerNA{}
    \item[] Justification: The paper does not release high-risk generative models, scraped datasets, language models, or systems with obvious misuse risk. The proposed method is a reconstruction training controller evaluated on public CT object data.

\item {\bf Licenses for existing assets}
    \item[] Question: Are the creators or original owners of assets (e.g., code, data, models), used in the paper, properly credited and are the license and terms of use explicitly mentioned and properly respected?
    \item[] Answer: \answerYes{}
    \item[] Justification: The paper cites the FIPS dataset and prior reconstruction methods used as baselines. Any released supplementary code or data should preserve the corresponding licenses and attribution requirements.

\item {\bf New assets}
    \item[] Question: Are new assets introduced in the paper well documented and is the documentation provided alongside the assets?
    \item[] Answer: \answerNA{}
    \item[] Justification: The paper does not introduce a new dataset or benchmark asset. The contribution is an algorithmic training-control framework evaluated on existing public data.

\item {\bf Crowdsourcing and research with human subjects}
    \item[] Question: For crowdsourcing experiments and research with human subjects, does the paper include the full text of instructions given to participants and screenshots, if applicable, as well as details about compensation (if any)?
    \item[] Answer: \answerNA{}
    \item[] Justification: The paper does not involve crowdsourcing, user studies, or experiments with human subjects.

\item {\bf Institutional review board (IRB) approvals or equivalent for research with human subjects}
    \item[] Question: Does the paper describe potential risks incurred by study participants, whether such risks were disclosed to the subjects, and whether Institutional Review Board (IRB) approvals (or an equivalent approval/review based on the requirements of your country or institution) were obtained?
    \item[] Answer: \answerNA{}
    \item[] Justification: The experiments use public non-human CT object data and do not involve human subjects, so IRB approval is not applicable.

\item {\bf Declaration of LLM usage}
    \item[] Question: Does the paper describe the usage of LLMs if it is an important, original, or non-standard component of the core methods in this research? Note that if the LLM is used only for writing, editing, or formatting purposes and does \emph{not} impact the core methodology, scientific rigor, or originality of the research, declaration is not required.
    \item[] Answer: \answerNA{}
    \item[] Justification: The core method development, experiments, and scientific claims do not rely on LLMs as a methodological component. Any use of LLMs for writing or editing does not affect the core methodology.

\end{enumerate}